\pdfoutput=1
\documentclass[11pt]{article}

\usepackage[final]{acl}

\usepackage{times}
\usepackage{latexsym}

\usepackage[algo2e]{algorithm2e} 

\usepackage{graphicx}
\usepackage{enumitem}
\usepackage{rotating}
\usepackage{multirow}                 % 一般用以设计表格，将所行合并
\usepackage{multicol}                 % 合并多列
\usepackage{multirow}                % 合并多行
\usepackage{float}   
\usepackage{makecell}                 % 三线表-竖线
\usepackage{booktabs}                 % 三线表-短细横线
\usepackage{times}  % DO NOT CHANGE THIS
\usepackage{helvet}  % DO NOT CHANGE THIS
\usepackage{courier}  % DO NOT CHANGE THIS
\usepackage{epstopdf}
\usepackage{subfiles}
\usepackage{amssymb} % 用于提供 \square 命令
\usepackage{tasks}
\usepackage{soul}
\usepackage[utf8]{inputenc}
\usepackage{amsmath}
\usepackage{amsthm}
\usepackage[switch]{lineno}
\usepackage{xspace}
\usepackage{array}
\usepackage{amssymb}
\usepackage{xcolor}
\usepackage{amsmath}
\usepackage{subcaption}
\usepackage{pifont}
\usepackage{booktabs}  % 提供三线表的支持
\usepackage{multirow}  % 提供合并单元格的支持
\usepackage{tabularx}  % 引入 tabularx 宏包
\usepackage{lipsum}    % 提供示例文本
\usepackage{adjustbox}  % 引入 adjustbox 宏包
\usepackage{algorithm}
\usepackage{algpseudocode}
\usepackage{float} % 加载 float 宏包

\restylefloat{algorithm}

\newcommand{\shortname}{SeAdpra}

\usepackage{newfloat}
\usepackage{listings}
\usepackage[T1]{fontenc}
\usepackage[utf8]{inputenc}

\usepackage{microtype}

\usepackage{inconsolata}

\usepackage{graphicx}

\title{Self-supervised Attribute-aware Dynamic Preference Ranking Alignment}

\author{First Author \\
  Affiliation / Address line 1 \\
  Affiliation / Address line 2 \\
  Affiliation / Address line 3 \\
  \texttt{email@domain} \\\And
  Second Author \\
  Affiliation / Address line 1 \\
  Affiliation / Address line 2 \\
  Affiliation / Address line 3 \\
  \texttt{email@domain} \\}

\author{
 \textbf{Hongyu Yang\textsuperscript{1}},
 \textbf{Qi Zhao\textsuperscript{1}},
 \textbf{Zhenhua Hu\textsuperscript{1}},
 \textbf{Rui Li\textsuperscript{1}},
\\
\\
 \textsuperscript{1}University of Science and Technology of China
\\
 \small{
   \href{mailto:email@domain}{hongyuyang@mail.ustc.edu.cn}
 }
}

\begin{document}
\maketitle
\begin{abstract}
Reinforcement Learning from Human Feedback and its variants excel in aligning with human intentions to generate helpful, harmless, and honest responses. 
However, most of them rely on costly human-annotated pairwise comparisons for supervised alignment, which is not suitable for list-level scenarios, such as community question answering.
Additionally, human preferences are influenced by multiple intrinsic factors in responses, leading to decision-making inconsistencies.
Therefore, we propose \textbf{Se}lf-supervised \textbf{A}ttribute-aware \textbf{d}ynamic \textbf{p}reference \textbf{ra}nking, called \shortname. \
It quantifies preference differences between responses based on Attribute-Perceptual Distance Factors (APDF) and dynamically determines the list-wise alignment order.
Furthermore, it achieves fine-grained preference difference learning and enables precise alignment with the optimal one.
We specifically constructed a challenging code preference dataset named StaCoCoQA, and introduced more cost-effective and scalable preference evaluation metrics: PrefHit and PrefRecall.
Extensive experimental results show that SeAdpra exhibits superior performance and generalizability on both StaCoCoQA and preference datasets from eight popular domains.

\end{abstract}

\section{Introduction}
\begin{figure}
    \centering
    \includegraphics[width=\linewidth]{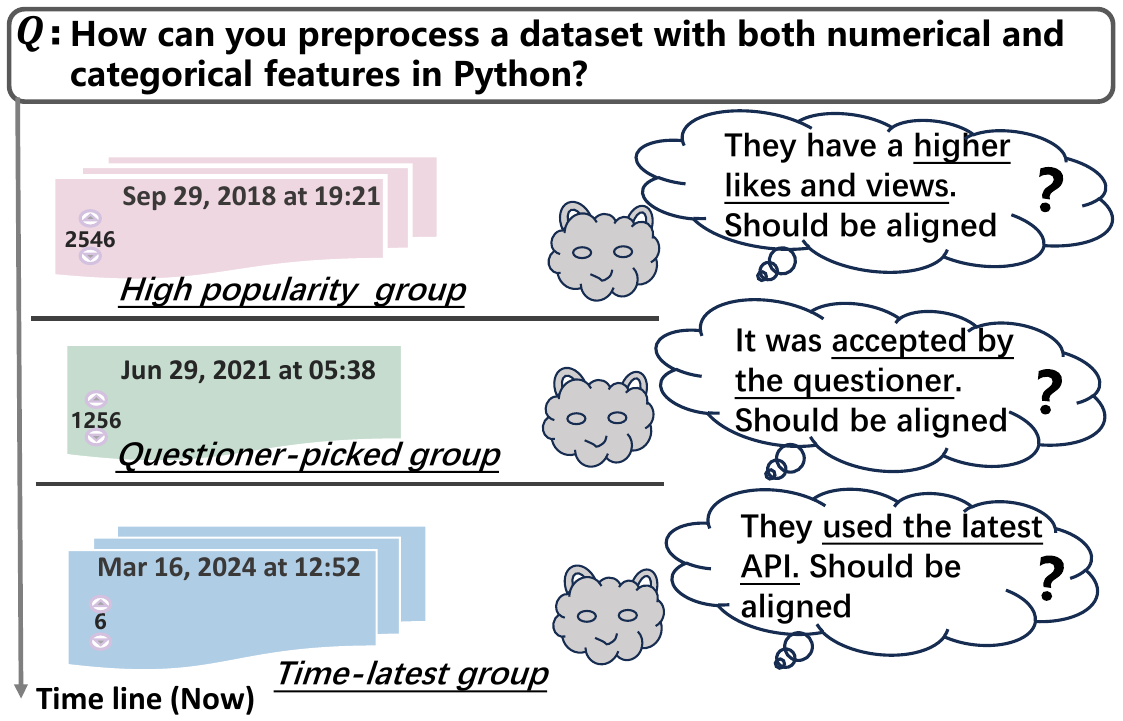}
    \caption{Which response should the LLMs align with? In the code community, each response has different attributes such as semantics, popularity, and timeliness, leading to potentially different optimal responses.}
    \label{intro}
        % \vspace{-0.2cm}
\end{figure}
Community Question Answering (CoQA) \cite{romeo2018flexible,wu2018question} seeks to generate responses that are semantically accurate and 
match the preferences of community members.
Currently, Reinforcement Learning from Human (or AI) Feedback (RLHF/RLAIF) \cite{christiano2017deep, bai2022constitutional} has enabled precise control of large language models (LLMs) for generating human-like responses \cite{stiennon2020learning, ouyang2022training}. 
However, applying it to CoQA remains underexplored.
Moreover, human preferences do not always follow a singular, value-based hierarchy. 
Decision-making can be influenced by various factors and may exhibit inconsistencies \cite{tversky1969intransitivity}, which undoubtedly presents a challenge for aligning LLMs with CoQA.

Existing methods are limited to pairwise comparison (one chosen and one rejected), such as reward model-based RLHF \cite{ouyang2022training}, offline supervised Direct Preference Optimization (DPO) \cite{rafailov2024direct}, as well as other variants like SLiC \cite{zhao2023slic} and pseudo-list RRHF \cite{yuan2024rrhf} that adopt pairwise hinge loss.
However, a real-world prompt may have multiple high-quality responses \cite{cui2023ultrafeedback}.
For example, in the coding community, the optimal one may vary with thier different attributes, such as semantics, popularity, and timeliness, as illustrated in Figure~\ref{intro}. 
Recently, some alignment methods have attempted to rank multiple preferred candidates.
PRO \cite{song2024preference} introduces a list-level maximum likelihood estimation loss to shift towards preference ranking but overlooks the attributes of responses. 
LiPO \cite{liu2024lipo} directly optimizes list-based ranking preferences and begins to address response labels, but has not yet addressed the integration of multiple labels.
Moreover, these supervised learning methods depend on human or AI annotations of preference pairs or lists to specify the best responses for alignment. However, preference data are relatively scarce and expensive to collect in practice \cite{casper2023open}.

% Earlier, RLHF \cite{ouyang2022training,stiennon2020learning,christiano2017deep} fitted a reward model to human preference datasets and then used Proximal Policy Optimization (PPO) \cite{schulman2017proximal} to optimize the language model policy (\(\pi_{Ref}\)) to generate responses that receive high rewards. To circumvent the complexities of the RLHF, the popular offline supervised Direct Preference Optimization (DPO) \cite{rafailov2024direct}, under the pair-wise Bradley-Terry assumption \cite{bradley1952rank} or list-wise Plackett-Luce model \cite{luce1959individual}, replaced rewards with the relative logistic probabilities of responses; the higher the relative log probability, the greater the preference.
% Parallel to this, SLiC \cite{zhao2023slic} and pseudo-list RRHF \cite{yuan2024rrhf} both use pair-wise hinge loss to align policy responses.
% These methods\cite{liu2024lipo,azar2024general,guo2024direct} are limited to single pair-wise responses for each prompt (one chosen and one rejected). 
% However, there might be multiple high-quality responses for a single prompt \cite{cui2023ultrafeedback}.
% To better align LLMs with CoQA
To address the above issues, we propose \shortname, a \textbf{Se}lf-supervised \textbf{a}ttribute-aware \textbf{d}ynamic \textbf{p}reference \textbf{ra}nking framework. It consists of three stages.
% It focuses on the intrinsic attributes through Attribute-Perceptual Distance Factors (APDF), quantifying the preference-level differences among multiple responses,
First, the Multi-Attribute Perception quantifies preference-level differences through Attribute-Perceptual Distance Factors (APDF), enabling the integration of multiple attributes for self-supervised dynamic ranking.
Second, the Perception Alignment aims to quickly adapts to domain knowledge and achieve precise alignment by aligning the optimal.
Third, the Perceptual Comparison performs multiple iterative comparisons on all candidates to learn on-chain preference differences. 
% In each iteration, it maximizes the reward for the optimal response and minimizes the penalty for the remaining responses based on preference levels.

% To accommodate improvements to our method and the expansion of comparison benchmarks, while 

For enhancing the cost-efficiency and domain applicability of the preference evaluation scheme, we propose new metrics that follow the 'CSTC' criterion (details in Appendix \ref{sec::cstc}), as an alternative to the costly win rate \cite{dudik2015contextual}, namely PrefHit and PrefRecall.
They can accommodate the expansion of benchmarks.
Aiming to validate the effectiveness of SeAdpra in specific domains, we have constructed a programming CoQA preference dataset, called StaCoCoQA, which contains over 60,738 programming directories and 9,978,474 entries. 
Our main contributions are as follows:

% \noindent (1) We introduce the Attribute Perceptual Distance Factor (APDF) to gauge the in preference-level gaps of multiple responses, replacing the binary judgment of preferred versus non-preferred. We propose an self-supervised dynamic preference ranking framework that achieves label-free list-wise preference alignment.

% \noindent (2) We present the StaCoCoQA, a large-scale, high-quality, real-time (as of May 2024) dataset for preference alignment in programming CoQA, and develop two new alignment metrics abided by the 'CSTC' criterion.

% \noindent (3) We conducted extensive experiments on eight hot public datasets and StaCoCoQA, providing a reference benchmark. 
% Along with the increase in safety, the experimental results demonstrate the superiority of the alignment of \shortname.
% \item We conducted extensive experiments on eight public datasets (such as Security, Chemistry, Cooking, Academia, etc.) and StaCoCoQA, providing a reference benchmark. Along with the increase in safety, the experimental results demonstrate the superiority of the preference alignment of \shortname.

\begin{itemize}[leftmargin=*]
\item We introduce the Attribute Perceptual Distance Factor (APDF) to gauge the in preference-level gaps of multiple responses, replacing the binary judgment of preferred versus non-preferred. We propose an self-supervised dynamic preference ranking framework that achieves label-free list-wise preference alignment.
\item We present the StaCoCoQA, a large-scale, high-quality, real-time (as of May 2024) dataset for preference alignment in programming CoQA, and develop two new alignment metrics abided by the 'CSTC' criterion.
\item We conducted extensive experiments on eight hot public datasets and StaCoCoQA, providing a reference benchmark. 
The experimental results demonstrate that \shortname \ excels in alignment while maintaining safety. 
% Additionally, we analyzed the effectiveness of the new metrics from multiple perspectives.
% \item We conducted extensive experiments on eight public datasets (such as Security, Chemistry, Cooking, Academia, etc.) and StaCoCoQA, providing a reference benchmark. Along with the increase in safety, the experimental results demonstrate the superiority of the preference alignment of \shortname.
\end{itemize}
\begin{figure}[t]
    \centering
    \includegraphics[width=\linewidth]{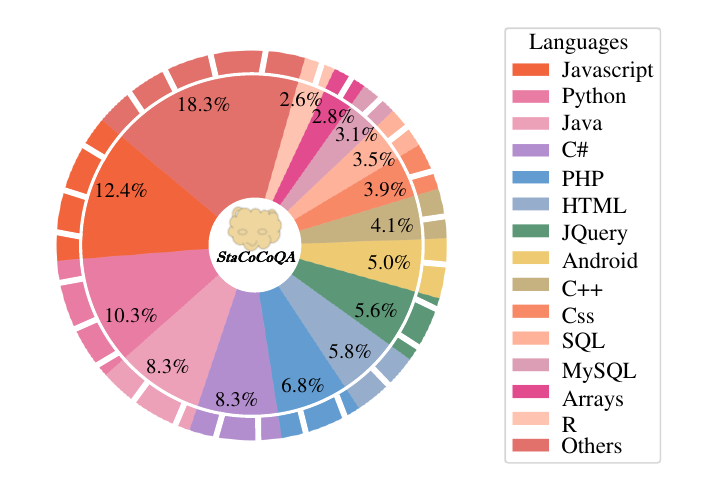}
    \caption{Showcasing the top-15 primary programming language categories in StaCoCoQA.}
    \label{fig:dataset}
    \vspace{-0.2cm}
\end{figure}

\section{Method}
% \begin{figure*}
%     \centering
%     \includegraphics[width=\linewidth]{latex/pic/framework1.pdf}
%     \caption{}
%     \label{intro}
% \end{figure*}
\begin{figure*}[t]
    \centering
    \includegraphics[width=\linewidth]{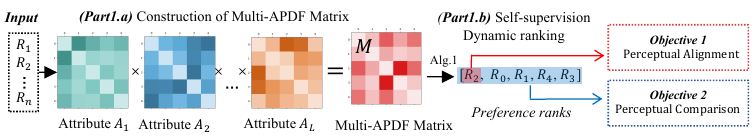}
    \caption{The overall framework of \shortname \
    , which includes: (Part1.) Multi-attribute Perception for quantifying preference, containing the Construction of Multi-APDF Matrix and Self-supervised dynamic ranking; (Part2.) Perceptual Alignment for aligning the optimal ranks objective; (Part3.) Perceptual Comparison on all candidates for learning on-chain preference difference.}
    \label{overall}
    \vspace{-0.5cm}
\end{figure*}
\begin{figure}[ht]
    \centering
    \includegraphics[width=\linewidth]{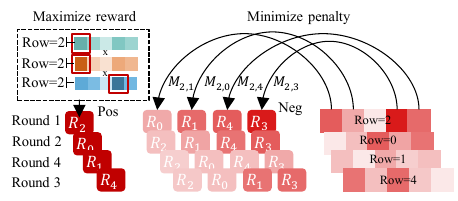}
    \caption{Implementation Workflow of Perceptual Comparison. In each round, the reward of the current positive is maximized, and the penalty for the remaining negative is minimized sequentially.}
    \vspace{-0.2cm}
\end{figure}
\subsection{Problem Definition}
Our goal is to align an LLM with user preferences in CoQA using our Unsupervised Attribute-aware Dynamic Preference Ranking strategy.
The training dataset is denoted as \(\mathcal{D} = \{Q^{i}, R^{i}\}_{i=1}^{N}\). For a given question \(Q\), it corresponds to a series of responses \(R = \{R_1, \ldots, R_M\}\), where each response \(R_i = (C, A)\), with \(C\) representing the content and \(A\) representing the scalable attributes. The size \(L\) of the scalable attribute \(A = \{A_1, \ldots, A_L\}\) is determined by community characteristics. For example, in the code community, \(L=3\) and \(A = \{S, P, T\}\).
Here, \(S\) represents the semantic similarity between \(C\) and \(Q\); \(P\) represents the popularity of \(R\), and \(T\) represents the creation time of each response.
% It is worth noting that for any \(i > j\), it is not necessarily true that \(R_i \succ R_j\), where \(\mathcal{\succ}\) denotes the degree of preference.
\subsection{Multi-attribute Perception}
% In this part, we will sequentially introduce the (1) Attribute-Perceptual Distance Factor (APDF), (2) the Construction of the Multi-APDF Matrix, and (3) the Self-supervision Dynamic Ranking.
% In this part, we sequentially introduce (1) Multi-attribute Perception, including the introduction of Attribute Perceptual Distance Factors for quantifying preference levels and the unsupervised dynamic ranking algorithm described in Algorithm~\ref{algo1}, (2) Perceptual Comparison for learning on-chain preference differences, and (3) Perceptual Alignment for aligning the best responses.

\subsubsection{Attribute-Perceptual Distance Factor}
The existing alignment optimization objectives \cite{rafailov2024direct, song2024preference} do not take into account the attributes of the candidates, which can differentiate their preferences. 
Therefore, there is a need to explore optimization methods that can effectively incorporate these attributes.
In this context, LambdARank \cite{burges2005learning, donmez2009local, wang2018lambdaloss, jagerman2022optimizing} introduces Lambda weights \(\lambda_{ij}\), which scale the gradient of each pair of scores based on the labels of the pairs to optimize a metric-driven loss function and effectively incorporating label information into the optimization process.

Inspired by the \(\lambda_{ij}\), the Attribute-Perceptual Distance Factor \(\delta_{i,j}\) is designed to quantify the preference difference between two candidates \(i\) and \(j\) in the optimization objective. It not only considers the positional relationship of candidates in preference ranks but also incorporates their label values through the gain function, and expressed as:
\begin{equation}
    \delta_{i,j} = (G(i) - G(j)) \cdot \left( T(i) - T(j) \right) 
\end{equation}
\begin{equation}
    T(i) = 1/log(l_i+1)
\end{equation}
where \(l_i\) and \(l_j\) are the ranking positions of response \(i\) and \(j\), respectively. 
The gain function \(G(\cdot)\) varies with different intrinsic attributes.

\subsubsection{Construction of the Multi-APDF Matrix}
Given the response \(R = \{R_1, \ldots, R_M\}\) to question \(Q\), the construction of the Multi-APDF matrix is a dot-product fusion of \(L\) Single-APDF matrix.
Based on the characteristics of the code community shown in Figure \ref{intro}, the main attributes that influence user preferences are semantics (text content), popularity, and creation time. 
% Additionally, the following phenomena occur: popularity accumulates over time, and there may be outdated information in responses due to updates and iterations of programming libraries.

\textbf{Semantic-APDF matrix} \(\Delta_{Se} = \{\delta_{Se_{ij}} | i,j \in M\}\), we define \(G^{Se}(i) = 2^{\varphi(i) - 1}\), where \(\varphi(i) = \cos(E_{Q}, E_{C_i})\). Here, \(E_{Q} \in \mathbb{R}^{q \times d}\) and \(E_{C_i} \in \mathbb{R}^{r \times d}\) represent the semantic vectors of the question \(Q\) and the text content \(C_i\) of response \(R_i\), encoded by prompt-based LLMs \cite{behnamghader2024llm2vec}. Here, \(q\) is the length of the question, \(r\) is the length of the text content, and \(d\) is the dimension of the embedding space.

\textbf{Popularity-APDF matrix} \(\Delta_{Po} = \{\delta_{Po_{ij}} | i,j \in M\}\)
To mitigate the bias caused by the accumulation of popularity over time, we apply time decay to \(P\) based on \(T\), denoted as \(\tilde{P}\). 
To avoid bias caused by extreme values and excessive numerical differences, we set \(G^{Po}(i) = \lg(\tilde{P_i} + 1)\).

\textbf{Multi-APDF matrix} on the scalable attribute \( A = \{ A_1, \ldots, A_L \} \) is represented generally as:
\begin{equation}
  \label{muapdf}
    \Delta_{M} =  {\textstyle \prod_{k=1}^{L}}  \Delta_{A_k}
\end{equation}
where \( \Delta_{A_k} \) is the APDF matrix corresponding to attribute \( A_k \). Similarly, The code Multi-APDF matrix \(\Delta_{M}^{code} \in \mathbb{R}^{M \times M}\) is represented as follows:
\begin{equation}
    \Delta_{M}^{code} = \Delta_{Se} \cdot \Delta_{Po}
\end{equation}
% we define the Semantic Perceptual Distance Factor matrix \(\Delta_{Se} = \{\delta_{Se_{ij}} | i,j \in M\}\) and the Popularity Perceptual Distance Factor matrix \(\Delta_{Po} = \{\delta_{Po_{ij}} | i,j \in M\}\).

% \begin{equation}
%   \label{muapdf}
%     \Delta_{MuAPDF} =  \prod_{k=1}^{L} \Delta_{A_k}
% \end{equation}
% To determine the optimal response order in Perceptual Comparison, in each iteration, we sequentially identify the response corresponding to the largest \( \delta_{MuAPDF}\).

\subsubsection{Self-supervision Dynamic Ranking}
To avoid relying on manually labeled alignment targets, we propose the Self-supervised Dynamic Ranking based on the Multi-APDF Matrix.  
It iteratively selects the most significant pair-wise distance (Multi-APDF \(\delta_{M}\)) and ranks the candidates according to the semantic ranks, which ensures that the ranking not only reflects pair-wise perceptual differences but also adheres to semantic priorities.
Its implementation details are provided in the Algorithm~\ref{algo1}.
The \(D^R\) represents the set of candidates' positions after dynamic ranking:
\begin{equation}
    D^R = \{i_1, i_2, \ldots, i_M\}
\end{equation}
\subsection{Perceptual Alignment}
Since the most effective learning for domain knowledge method is SFT \cite{stiennon2020learning}, and the most direct one in alignment is also to perform SFT on a high-quality preference dataset \cite{rafailov2024direct}, 
we align the optimal response by treating the first response in dynamic ranking as the target for SFT for the question \( Q \).
The first optimization objective is represented as follows:
\begin{equation}
   L_{Pa} = - \frac{1}{|R_b|} \sum_{j=1}^{|R_b|} \log P(R_b(j) | Q, R_b(<j))
\end{equation}
where \( R_{D^R(0)}\) denotes as \( R_b \). The \( D^R(i) \) is the \( i \)-th element, and \( R_b(j) \) is the \( j \)-th token.

\subsection{Perceptual Comparison}
In terms of many list-wise loss functions, the softmax cross-entropy loss in ListNet \cite{cao2007learning} uses double summation to emphasize comparisons between different samples, making it suitable for ranking loss. Therefore, we adopt it as the basis for the second optimization objective and conduct a total of \(M-1\) iterative comparisons.
To deepen the impact of preference differences, for each iteration, we maximize the reward for positive and minimize the penalty for remains negative sequentially.

\textbf{Maximizing the reward} is achieved by finding all maximum value in the mapped row of the alignment target in all Single-APDF matrix, and then multiplying the values together. For the \(m\)-th comparison, it is represented as follows:
\begin{equation}
  W_{m}^{r}= {\textstyle \prod_{k=1}^{L}} max(\Delta_{A_k}(D^R(m),\cdot))
\end{equation}
where \(\Delta_{A_k}(i, j)\) refers to the element at the \(i\)-th row and \(j\)-th column of \(\Delta_{A_k}\), and \(\cdot\) represents all elements in the row or column.

\textbf{Minimizing the penalty} involves differentiating the penalty strengths based on preference levels, where a slight penalty is applied to \( R_{D_R(i)} \) and a stronger penalty is applied to \( R_{D_R(i+1)} \). This approach contrasts with the existing method, which applies the same penalty to all negative examples, and ensures that the penalty for responses ranked higher in the self-supervised ranking \( D_R \) is minimized.
For the negative \( R_i \), its penalty is represented as follows:
\begin{equation}
\begin{aligned}
     W_{i}^{p}  =  sort(\Delta_{M}(D^R(m),\cdot ))(i)
\end{aligned}
\end{equation}
where \(sort(\cdot)\) is the function that sorts in an ascending order. 
\(\Delta_{M}(i, j)\) is the \(i\)-th row and the \(j\)-th APDF in the Multi-APDF matrix.

To achieve on-chain ranking and fine-grained distinction among all responses, unlike traditional optimization methods that sequentially remove the optimal response, all responses participate in each iteration. Moreover, the corresponding penalties or rewards for the responses change throughout the iterations.
The second optimization objective is represented as:
\begin{equation}
   L_{Pc} = -\sum_{m=1}^{M-1} \log \left( \frac{ \tau_r(b)}{\sum_{i \ne b}^{M} \tau_p(i) + \tau_r(b)} \right)
\end{equation}
\begin{equation}
    \tau_r (b)= \exp(\pi_s(Q, R_{b})) * W_{m}^{r}
\end{equation}
\begin{equation}
    \tau_p (i)= \exp(\pi_s(Q, R_i)) * W_{i}^{p}
\end{equation}
\begin{equation}
    \pi_s(Q, R_i) = \frac{1}{t} {\textstyle \sum_{k=1}^{t}\log P(r_k|Q, r_{<k}) }
\end{equation}
Here, \(D^R(m)\) denotes as \(b\). the \(\pi_s(\cdot)\) represents a policy network that replaces the reward in RLHF with language modeling logits. The labeled response \(R\), composed of \(t\) tokens, is denoted as \(R_i = \{r_1,\ldots,r_t\}\).
Finally, \shortname\ enables LLMs to be trained by the following objective:
\begin{equation}
    Loss= L_{Pc} + \alpha \cdot L_{Pa}
    \label{eq::final}
\end{equation}
To avoid overfitting the initial best response, \(\alpha\) will control the balance between it and the remaining preferences, thereby ensuring text quality.

\section{Experiments}

%%%%%%%%%%%%%%%%%%%%%%%%%%%%%%%%%%%%%%%%%%%%%%%%%%%%%%%%%%%%%%%%%%%%%%%%%%%%%%%%%%%%%%%%%%%%%%%%%%%%%%

%%%%%%%%%%%%%%%%%%%%%%%%%%%%%%%%%%%%%%%%%%%%%%
\begin{table*}[t]
\setlength{\tabcolsep}{3pt}
\centering
\renewcommand{\arraystretch}{1.1}
\tabcolsep=0.2cm
\begin{adjustbox}{max width=\textwidth}  % Set the maximum width to text width
\begin{tabular}{c| cccc ||  c| cc cc}
\toprule
General & \multicolumn{3}{c}{Preference} & Accuracy & Supervised & \multicolumn{3}{c}{Preference} & Accuracy \\ 
LLMs & PrefHit & PrefRecall & Reward & BLEU & Alignment & PrefHit & PrefRecall & Reward & BLEU \\ 
\midrule
GPT-J & 0.2572 & 0.6268 & 0.2410 & 0.0923 & Llama2-7B & 0.2029 & 0.803 & 0.0933 & 0.0947 \\
Pythia-2.8B & 0.3370 & 0.6449 & 0.1716 & 0.1355 & SFT & 0.2428 & 0.8125 & 0.1738 & 0.1364 \\
Qwen2-7B & 0.2790 & 0.8179 & 0.1593 & 0.2530 & Slic & 0.2464 & 0.6171 & 0.1700 & 0.1400 \\
Qwen2-57B & 0.3086 & 0.6481 & 0.6854 & 0.2568 & RRHF & 0.3297 & 0.8234 & 0.2263 & 0.1504 \\
Qwen2-72B & 0.3212 & 0.5555 & 0.6901 & 0.2286 & DPO-BT & 0.2500 & 0.8125 & 0.1728 & 0.1363 \\ 
StarCoder2-15B & 0.2464 & 0.6292 & 0.2962 & 0.1159 & DPO-PT & 0.2572 & 0.8067 & 0.1700 & 0.1348 \\
ChatGLM4-9B & 0.2246 & 0.6099 & 0.1686 & 0.1529 & PRO & 0.3025 & 0.6605 & 0.1802 & 0.1197 \\ 
Llama3-8B & 0.2826 & 0.6425 & 0.2458 & 0.1723 & \textbf{\shortname}* & \textbf{0.3659} & \textbf{0.8279} & \textbf{0.2301} & \textbf{0.1412} \\ 
\bottomrule
\end{tabular}
\end{adjustbox}
\caption{Main results on the StaCoCoQA. The left shows the performance of general LLMs, while the right presents the performance of the fine-tuned LLaMA2-7B across various strong benchmarks for preference alignment. Our method SeAdpra is highlighted in \textbf{bold}.}
\label{main}
\vspace{-0.2cm}
\end{table*}
%%%%%%%%%%%%%%%%%%%%%%%%%%%%%%%%%%%%%%%%%%%%%%%%%%%%%%%%%%%%%%%%%%%%%%%%%%%%%%%%%%%%%%%%%%%%%%%%%%%%
\begin{table}[h]
\centering
\renewcommand{\arraystretch}{1.02}
\begin{adjustbox}{width=0.48\textwidth} % Adjust table width
\begin{tabularx}{0.495\textwidth}{p{1.2cm} p{0.7cm} p{0.95cm}p{0.95cm}p{0.7cm}p{0.7cm}}
     \toprule
    \multirow{2}{*}{\small \textbf{Dataset}} & \multirow{2}{*}{\small Model} & \multicolumn{2}{c}{\small Preference} & \multicolumn{2}{c}{\small Acc } \\ 
    & & \small \textit{PrefHit} & \small \textit{PrefRec} & \small \textit{Reward} & \small \textit{Rouge} \\ 
    \midrule
    \multirow{2}{*}{\small \textbf{Academia}}   & \small PRO & 33.78 & 59.56 & 69.94 & 9.84 \\ 
                                & \small \textbf{Ours} & 36.44 & 60.89 & 70.17 & 10.69 \\ 
    \midrule
    \multirow{2}{*}{\small \textbf{Chemistry}}  & \small PRO & 36.31 & 63.39 & 69.15 & 11.16 \\ 
                                & \small \textbf{Ours} & 38.69 & 64.68 & 69.31 & 12.27 \\ 
    \midrule
    \multirow{2}{*}{\small \textbf{Cooking}}    & \small PRO & 35.29 & 58.32 & 69.87 & 12.13 \\ 
                                & \small \textbf{Ours} & 38.50 & 60.01 & 69.93 & 13.73 \\ 
    \midrule
    \multirow{2}{*}{\small \textbf{Math}}       & \small PRO & 30.00 & 56.50 & 69.06 & 13.50 \\ 
                                & \small \textbf{Ours} & 32.00 & 58.54 & 69.21 & 14.45 \\ 
    \midrule
    \multirow{2}{*}{\small \textbf{Music}}      & \small PRO & 34.33 & 60.22 & 70.29 & 13.05 \\ 
                                & \small \textbf{Ours} & 37.00 & 60.61 & 70.84 & 13.82 \\ 
    \midrule
    \multirow{2}{*}{\small \textbf{Politics}}   & \small PRO & 41.77 & 66.10 & 69.52 & 9.31 \\ 
                                & \small \textbf{Ours} & 42.19 & 66.03 & 69.74 & 9.38 \\ 
    \midrule
    \multirow{2}{*}{\small \textbf{Code}} & \small PRO & 26.00 & 51.13 & 69.17 & 12.44 \\ 
                                & \small \textbf{Ours} & 27.00 & 51.77 & 69.46 & 13.33 \\ 
    \midrule
    \multirow{2}{*}{\small \textbf{Security}}   & \small PRO & 23.62 & 49.23 & 70.13 & 10.63 \\ 
                                & \small \textbf{Ours} & 25.20 & 49.24 & 70.92 & 10.98 \\ 
    \midrule
    \multirow{2}{*}{\small \textbf{Mean}}       & \small PRO & 32.64 & 58.05 & 69.64 & 11.51 \\ 
                                & \small \textbf{Ours} & \textbf{34.25} & \textbf{58.98} & \textbf{69.88} & \textbf{12.33} \\ 
    \bottomrule
\end{tabularx}
\end{adjustbox}
\caption{Main results (\%) on eight publicly available and popular CoQA datasets, comparing the strong list-wise benchmark PRO and \textbf{ours with bold}.}
\label{public}
\end{table}

%%%%%%%%%%%%%%%%%%%%%%%%%%%%%%%%%%%%%%%%%%%%%%%%%%%%%
\begin{table}[h]
\centering
\renewcommand{\arraystretch}{1.02}
\begin{tabularx}{0.48\textwidth}{p{1.45cm} p{0.56cm} p{0.6cm} p{0.6cm} p{0.50cm} p{0.45cm} X}
\toprule
\multirow{2}{*}{Method} & \multicolumn{3}{c}{Preference \((\uparrow)\)} & \multicolumn{3}{c}{Accuracy \((\uparrow)\)} \\ \cmidrule{2-4} \cmidrule{5-7}
& \small PrefHit & \small PrefRec & \small Reward & \small CoSim & \small BLEU & \small Rouge \\ \midrule
\small{SeAdpra} & \textbf{34.8} & \textbf{82.5} & \textbf{22.3} & \textbf{69.1} & \textbf{17.4} & \textbf{21.8} \\ 
\small{-w/o PerAl} & \underline{30.4} & 83.0 & 18.7 & 68.8 & \underline{12.6} & 21.0 \\
\small{-w/o PerCo} & 32.6 & 82.3 & \underline{24.2} & 69.3 & 16.4 & 21.0 \\
\small{-w/o \(\Delta_{Se}\)} & 31.2 & 82.8 & 18.6 & 68.3 & \underline{12.4} & 20.9 \\
\small{-w/o \(\Delta_{Po}\)} & \underline{29.4} & 82.2 & 22.1 & 69.0 & 16.6 & 21.4 \\
\small{\(PerCo_{Se}\)} & 30.9 & 83.5 & 15.6 & 67.6 & \underline{9.9} & 19.6 \\
\small{\(PerCo_{Po}\)} & \underline{30.3} & 82.7 & 20.5 & 68.9 & 14.4 & 20.1 \\ 
\bottomrule
\end{tabularx}
\caption{Ablation Results (\%). \(PerCo_{Se}\) or \(PerCo_{Po}\) only employs Single-APDF in Perceptual Comparison, replacing \(\Delta_{M}\) with \(\Delta_{Se}\) or \(\Delta_{Po}\). The bold represents the overall effect. The underlining highlights the most significant metric for each component's impact.}
\label{ablation}
% \vspace{-0.2cm}
\end{table}

\subsection{Dataset}

% These CoQA datasets contain questions and answers from the Stack Overflow data dump\footnote{https://archive.org/details/stackexchange}, intended for training preference models. 

Due to the additional challenges that programming QA presents for LLMs and the lack of high-quality, authentic multi-answer code preference datasets, we turned to StackExchange \footnote{https://archive.org/details/stackexchange}, a platform with forums that are accompanied by rich question-answering metadata. Based on this, we constructed a large-scale programming QA dataset in real-time (as of May 2024), called StaCoCoQA. It contains over 60,738 programming directories, as shown in Table~\ref{tab:stacocoqa_tags}, and 9,978,474 entries, with partial data statistics displayed in Figure~\ref{fig:dataset}. The data format of StaCoCoQA is presented in Table~\ref{fig::stacocoqa}.

The initial dataset \(D_I\) contains 24,101,803 entries, and is processed by the following steps:
(1) Select entries with "Questioner-picked answer" pairs to represent the preferences of the questioners, resulting in 12,260,106 entries in the \(D_Q\).
(2) Select data where the question includes at least one code block to focus on specific-domain programming QA, resulting in 9,978,474 entries in the dataset \(D_C\).
(3) All HTML tags were cleaned using BeautifulSoup \footnote{https://beautiful-soup-4.readthedocs.io/en/latest/} to ensure that the model is not affected by overly complex and meaningless content.
(4) Control the quality of the dataset by considering factors such as the time the question was posted, the size of the response pool, the difference between the highest and lowest votes within a pool, the votes for each response, the token-level length of the question and the answers, which yields varying sizes: 3K, 8K, 18K, 29K, and 64K. 
The controlled creation time variable and the data details after each processing step are shown in Table~\ref{tab:statistics}.

To further validate the effectiveness of SeAdpra, we also select eight popular topic CoQA datasets\footnote{https://huggingface.co/datasets/HuggingFaceH4/stack-exchange-preferences}, which have been filtered to meet specific criteria for preference models \cite{askell2021general}. Their detailed data information is provided in Table~\ref{domain}.
\subsection{Evaluation Metrics}
\label{sec: metric}
For preference evaluation, we design PrefHit and PrefRecall, adhering to the "CSTC" criterion outlined in Appendix \ref{sec::cstc}, which overcome the limitations of existing evaluation methods, as detailed in Appendix \ref{metric::mot}.
In addition, we demonstrate the effectiveness of thees new evaluation from two main aspects: 1) consistency with traditional metrics, and 2) applicability in different application scenarios in Appendix \ref{metric::ana}.
Following the previous \cite{song2024preference}, we also employ a professional reward.
% Following the previous \cite{song2024preference}, we also employ a professional reward model\footnote{https://huggingface.co/OpenAssistant/reward-model-deberta-v3-large} \cite{song2024preference}, denoted as the Reward.

For accuracy evaluation, we alternately employ BLEU \cite{papineni2002bleu}, RougeL \cite{lin2004rouge}, and CoSim. Similar to codebertscore \cite{zhou2023codebertscore}, CoSim not only focuses on the semantics of the code but also considers structural matching.
Additionally, the implementation details of SeAdpra are described in detail in the Appendix \ref{sec::imp}.
\subsection{Main Results}
We compared the performance of \shortname with general LLMs and strong preference alignment benchmarks on the StaCoCoQA dataset, as shown in Table~\ref{main}. Additionally, we compared SeAdpra with the strongly supervised alignment model PRO \cite{song2024preference} on eight publicly available CoQA datasets, as presented in Table~\ref{public} and Figure~\ref{fig::public}.

\textbf{Larger Model Parameters, Higher Preference.}
Firstly, the Qwen2 series has adopted DPO \cite{rafailov2024direct} in post-training, resulting in a significant enhancement in Reward.
In a horizontal comparison, the performance of Qwen2-7B and LLaMA2-7B in terms of PrefHit is comparable.
Gradually increasing the parameter size of Qwen2 \cite{qwen2} and LLaMA leads to higher PrefHit and Reward.
Additionally, general LLMs continue to demonstrate strong capabilities of programming understanding and generation preference datasets, contributing to high BLEU scores.
These findings indicate that increasing parameter size can significantly improve alignment.

\textbf{List-wise Ranking Outperforms Pair-wise Comparison.}
Intuitively, list-wise DPO-PT surpasses pair-wise DPO-{BT} on PrefHit. Other list-wise methods, such as RRHF, PRO, and our \shortname, also undoubtedly surpass the pair-wise Slic.

\textbf{Both Parameter Size and Alignment Strategies are Effective.}
Compared to other models, Pythia-2.8B achieved impressive results with significantly fewer parameters .
Effective alignment strategies can balance the performance differences brought by parameter size. For example, LLaMA2-7B with PRO achieves results close to Qwen2-57B in PrefHit. Moreover, LLaMA2-7B combined with our method SeAdpra has already far exceeded the PrefHit of Qwen2-57B.

\textbf{Rather not Higher Reward, Higher PrefHit.}
It is evident that Reward and PrefHit are not always positively correlated, indicating that models do not always accurately learn human preferences and cannot fully replace real human evaluation. Therefore, relying solely on a single public reward model is not sufficiently comprehensive when assessing preference alignment.

% In conclusion, during ensuring precise alignment, SeAdpra will focuse on PrefHit@1, even though the trade-off between PrefHit and PrefRecall is a common issue and increasing recall may sometimes lead to a decrease in hit rate. The positive correlation between Reward and BLEU, indicates that improving the quality of the generated text typically enhances the Reward. 
% Most importantly, evaluating preferences solely based on reward is clearly insufficient, as a high reward does not necessarily correspond to a high PrefHit or PrefRecall.
%%%%%%%%%%%%%%%%%%%%%%%%%%%%%%%%%%%%%%%%%%%
%%%%%%%%%%%%
\begin{figure}
  \centering
  \begin{subfigure}{0.49\linewidth}
    \includegraphics[width=\linewidth]{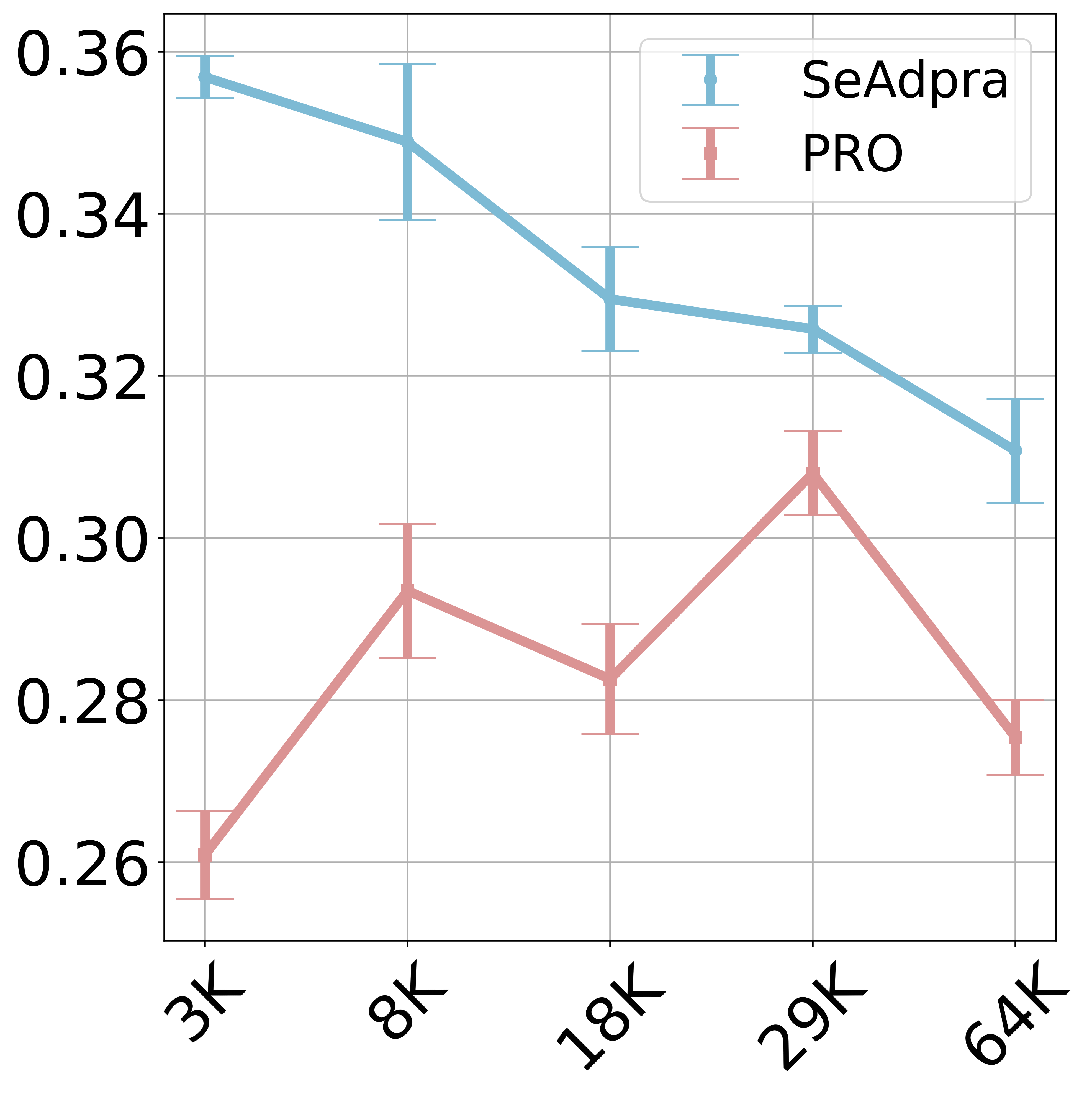}
    \caption{The PrefHit}
    \label{scale:hit}
  \end{subfigure}
  \begin{subfigure}{0.49\linewidth}
    \includegraphics[width=\linewidth]{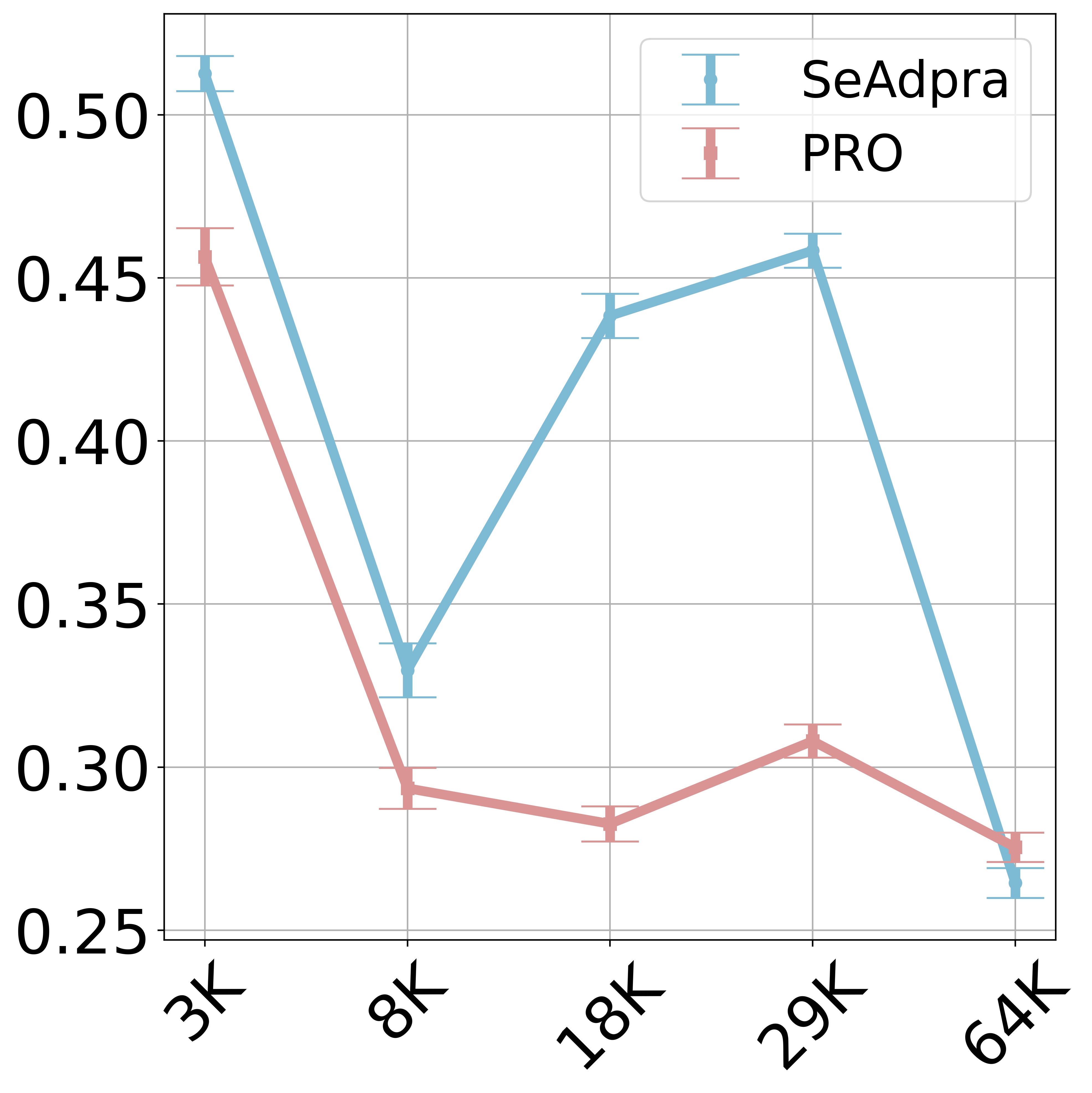}
    \caption{The PrefRecall}
    \label{scale:recall}
  \end{subfigure}
  \medskip
  \begin{subfigure}{0.48\linewidth}
    \includegraphics[width=\linewidth]{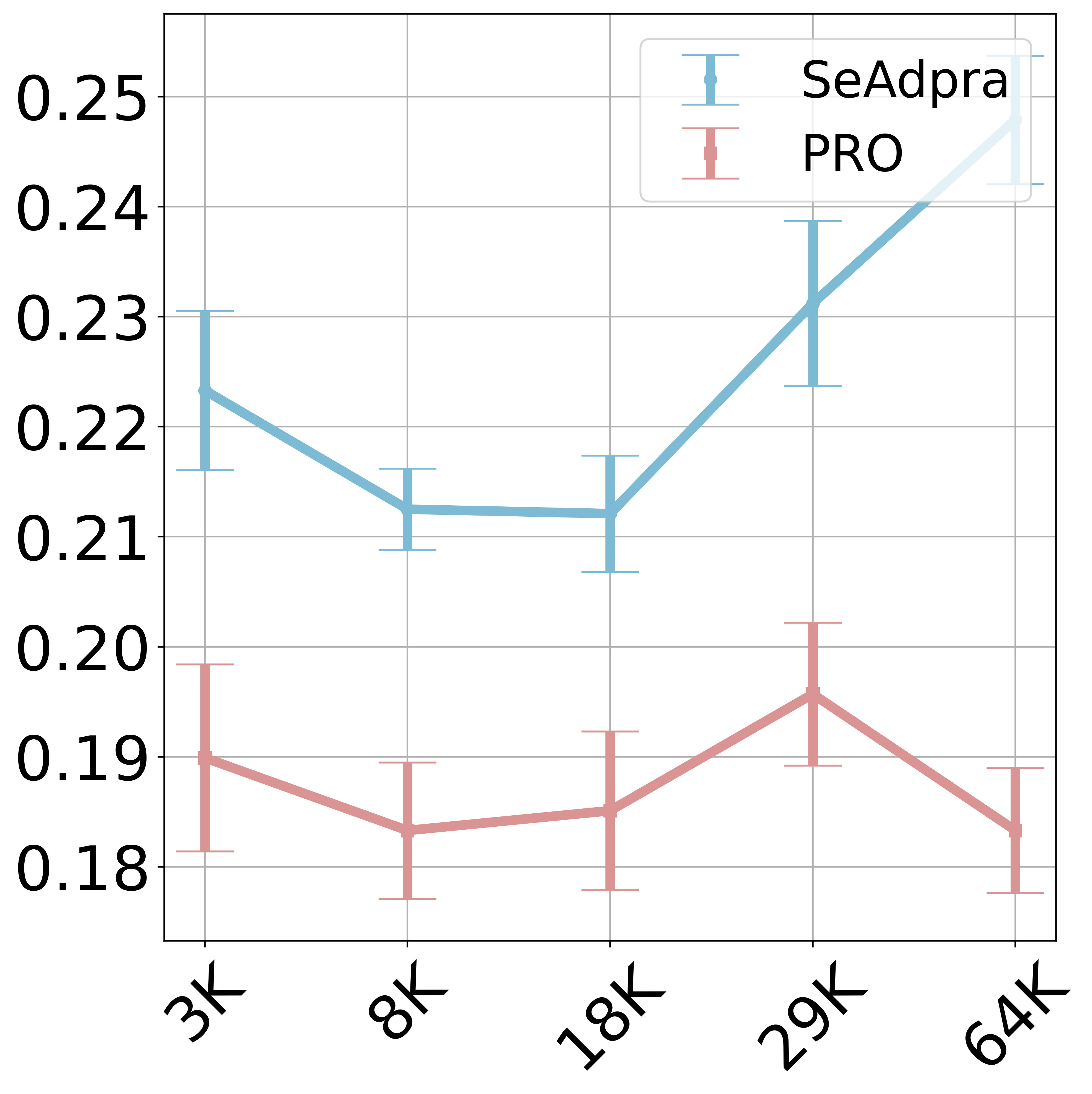}
    \caption{The Reward}
    \label{scale:reward}
  \end{subfigure}
  \begin{subfigure}{0.48\linewidth}
    \includegraphics[width=\linewidth]{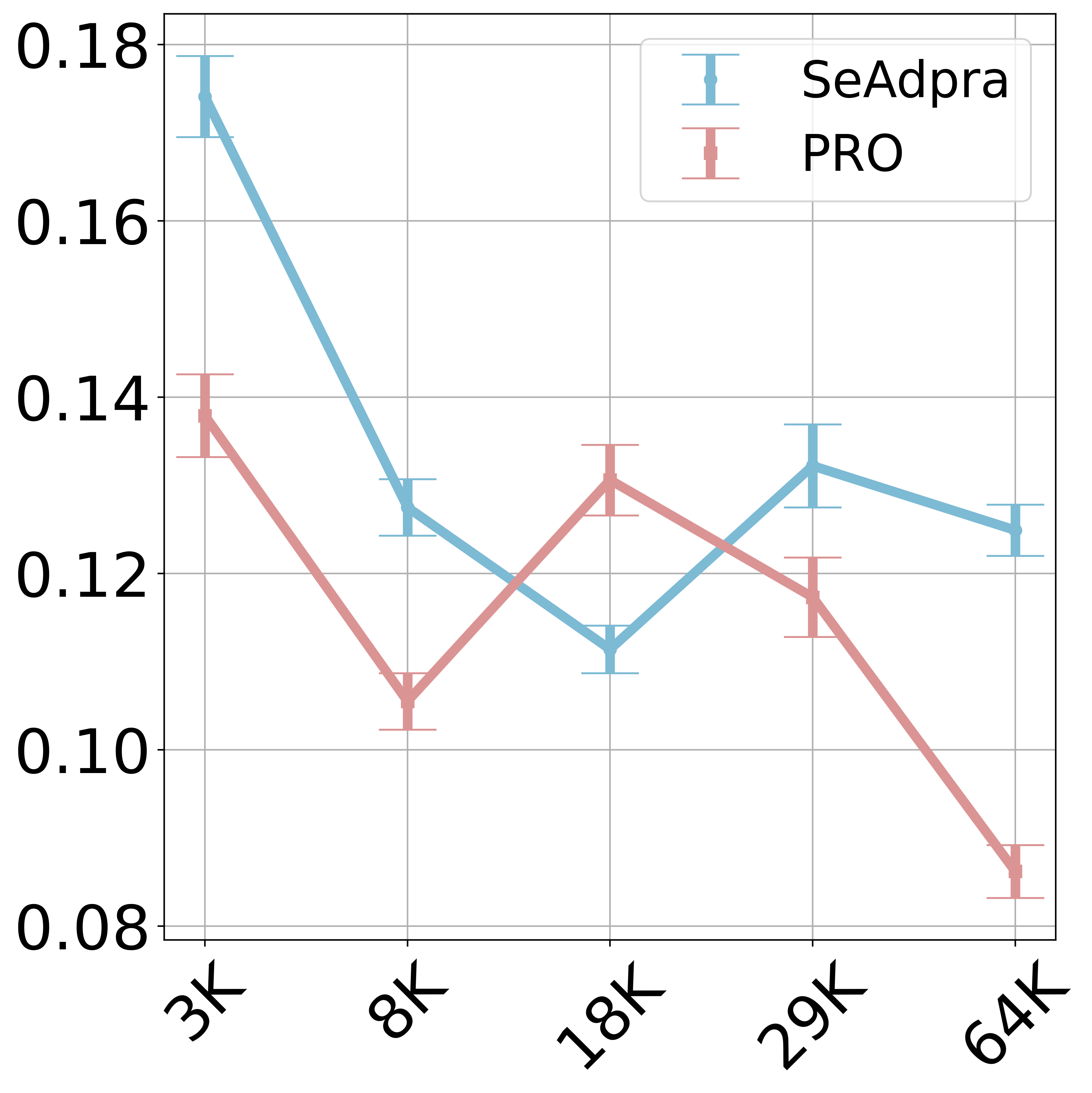}
    \caption{The BLEU}
    \label{scale:bleu}
  \end{subfigure}
  \caption{The performance with Confidence Interval (CI) of our SeAdpra and PRO at different data scales.}
  \label{fig:scale}
  % \vspace{-0.2cm}
\end{figure}
%%%%%%%%%%%%%%%%%%%%%%%%%%%%%%%%%%%%%%%%%%%%%%%%%%%%%%%%%%%%%%%%%%%%%%%%%%%%%%%%%%%%%%%%%%%%%%%%%%%%%%%%%%%%%%%%

\subsection{Ablation Study}

In this section, we discuss the effectiveness of each component of SeAdpra and its impact on various metrics. The results are presented in Table \ref{ablation}.

\textbf{Perceptual Comparison} aims to prevent the model from relying solely on linguistic probability ordering while neglecting the significance of APDF. Removing this Reward will significantly increase the margin, but PrefHit will decrease, which may hinder the model's ability to compare and learn the preference differences between responses.

\textbf{Perceptual Alignment} seeks to align with the optimal responses; removing it will lead to a significant decrease in PrefHit, while the Reward and accuracy metrics like CoSim will significantly increase, as it tends to favor preference over accuracy.

\textbf{Semantic Perceptual Distance} plays a crucial role in maintaining semantic accuracy in alignment learning. Removing it leads to a significant decrease in BLEU and Rouge. Since sacrificing accuracy recalls more possibilities, PrefHit decreases while PrefRecall increases. Moreover, eliminating both Semantic Perceptual Distance and Perceptual Alignment in \(PerCo_{Po}\) further increases PrefRecall, while the other metrics decline again, consistent with previous observations.

\textbf{Popularity Perceptual Distance} is most closely associated with PrefHit. Eliminating it causes PrefHit to drop to its lowest value, indicating that the popularity attribute is an extremely important factor in code communities.

% In summary, each module has a varying impact on preference and accuracy, but all outperform their respective foundation models and other baselines, as shown in Table \ref{main}, proving their effectiveness.

\subsection{Analysis and Discussion}

\textbf{SeAdpra adept at high-quality data rather than large-scale data.}
In StaCoCoQA, we tested PRO and SeAdpra across different data scales, and the results are shown in Figure~\ref{fig:scale}.
Since we rely on the popularity and clarity of questions and answers to filter data, a larger data scale often results in more pronounced deterioration in data quality. In Figure~\ref{scale:hit}, SeAdpra is highly sensitive to data quality in PrefHit, whereas PRO demonstrates improved performance with larger-scale data. Their performance on Prefrecall is consistent. In the native reward model of PRO, as depicted in Figure~\ref{scale:reward}, the reward fluctuations are minimal, while SeAdpra shows remarkable improvement.

\textbf{SeAdpra is relatively insensitive to ranking length.} 
We assessed SeAdpra's performance on different ranking lengths, as shown in Figure 6a. Unlike PRO, which varied with increasing ranking length, SeAdpra shows no significant differences across different lengths. There is a slight increase in performance on PrefHit and PrefRecall. Additionally, SeAdpra performs better at odd lengths compared to even lengths, which is an interesting phenomenon warranting further investigation.

\begin{figure}
  \centering
  \begin{subfigure}{0.475\linewidth}
    \includegraphics[width=\linewidth]{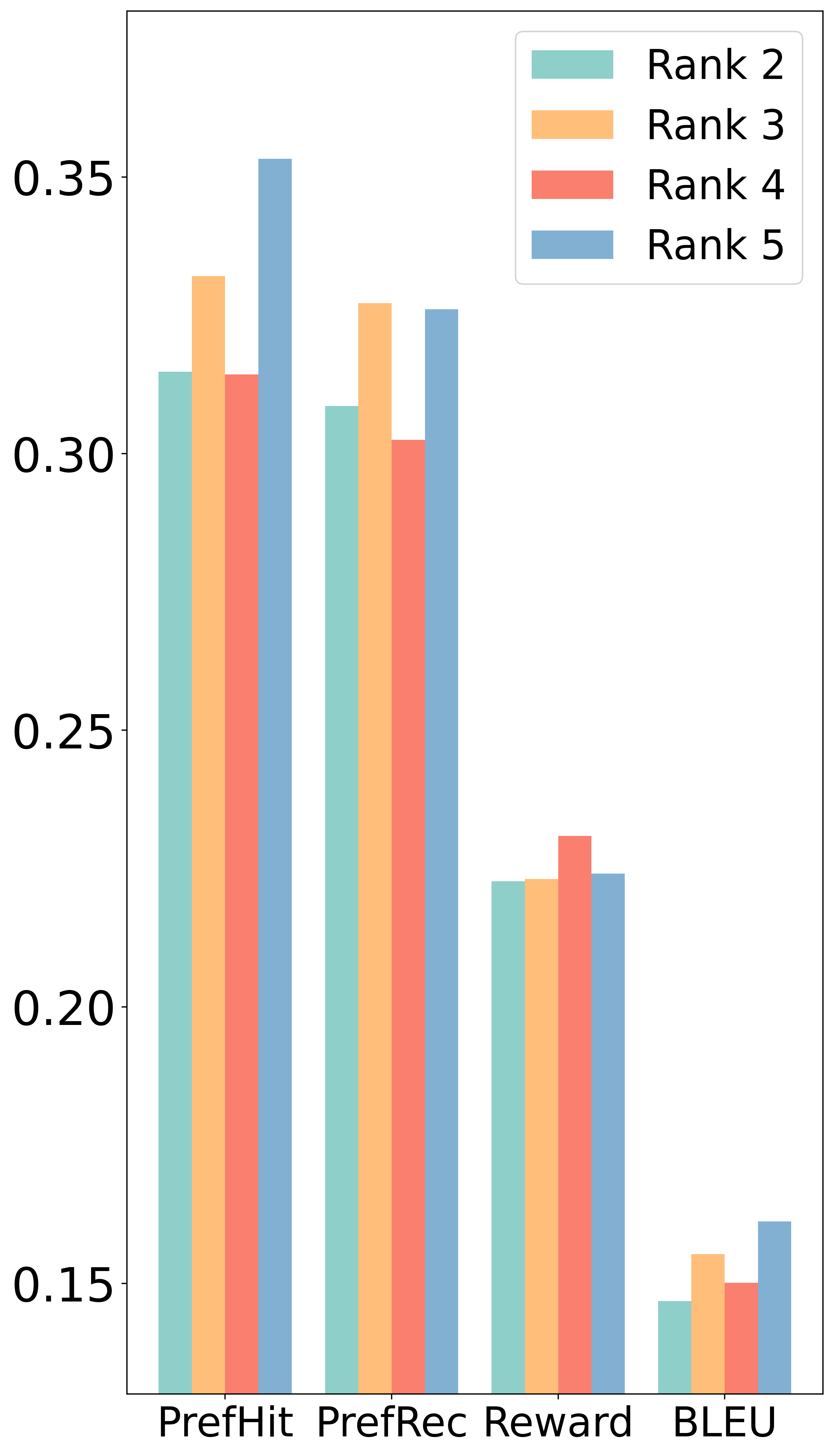}
    \caption{Ranking length}
    \label{para:rank}
  \end{subfigure}
  \begin{subfigure}{0.475\linewidth}
    \includegraphics[width=\linewidth]{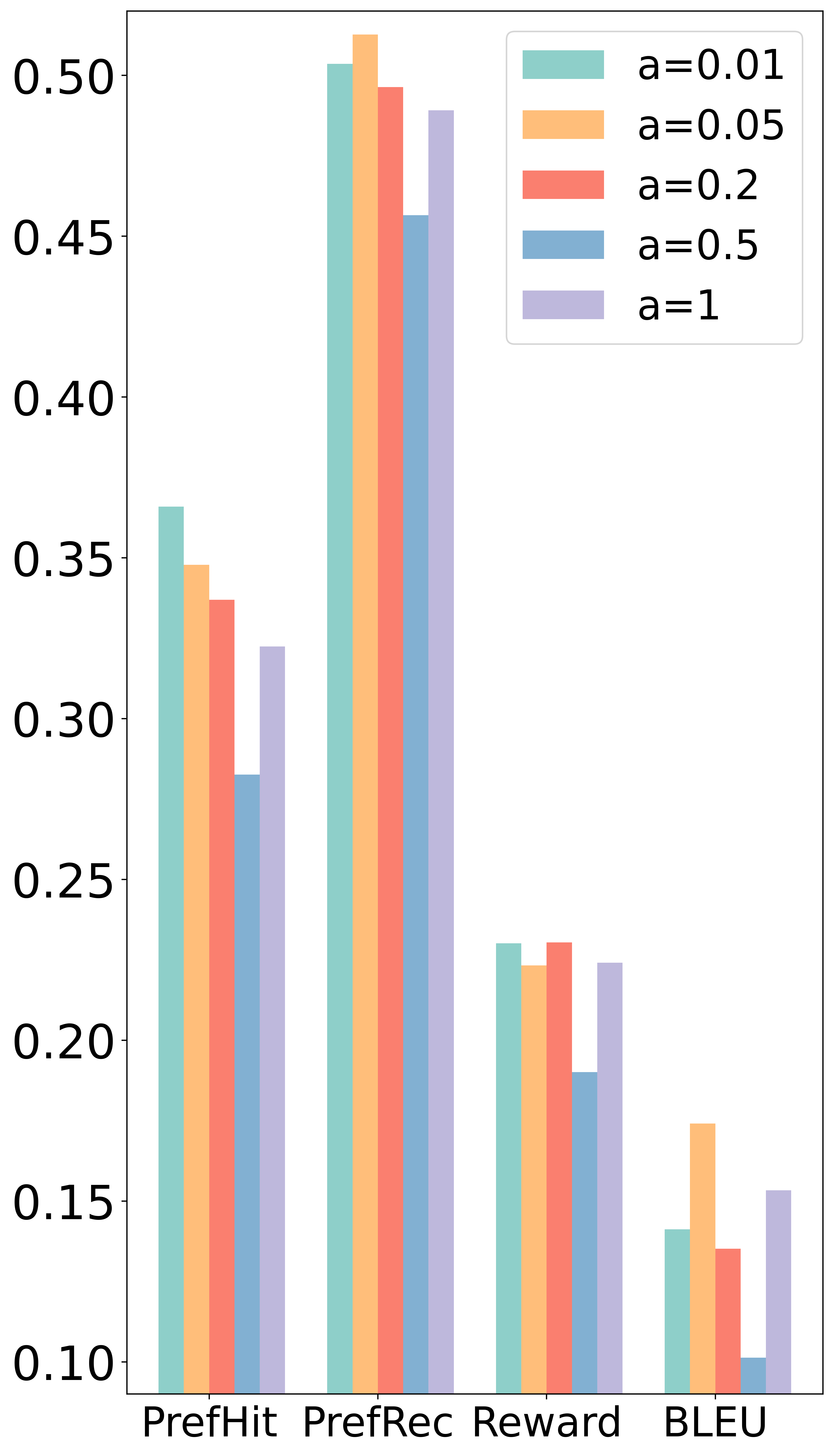}
    \caption{The \(\alpha\) in \(Loss\)}
    \label{para:weight}
  \end{subfigure}
  \caption{Parameters Analysis. Results of experiments on different ranking lengths and the weight \(\alpha\) in \(Loss\).}
  \label{fig:para}
  % \vspace{-0.2cm}
\end{figure}

\textbf{Balance Preference and Accuracy.} 
We analyzed the effect of control weights for Perceptual Comparisons in the optimization objective on preference and accuracy, with the findings presented in Figure~\ref{para:weight}.
When \( \alpha \) is greater than 0.05, the trends in PrefHit and BLEU are consistent, indicating that preference and accuracy can be optimized in tandem. However, when \( \alpha \) is 0.01, PrefHit is highest, but BLEU drops sharply.
Additionally, as \( \alpha \) changes, the variations in PrefHit and Reward, which are related to preference, are consistent with each other, reflecting their unified relationship in the optimization. Similarly, the variations in Recall and BLEU, which are related to accuracy, are also consistent, indicating a strong correlation between generation quality and comprehensiveness. 

%%%%%%%%%%%%%%%%%%%%%%%%%%%%%%%%%%%%%%%%%%%%%%%%%%%%%%%%%%%%%%%%%%%%%%%%%%%%%%%%%
%%%%%%%%%%%%%%%%%%%%%%%%%%%%%%%%%%%%%%%%%%%%
\begin{figure*}
  \centering
  \begin{subfigure}{0.305\linewidth}
    \includegraphics[width=\linewidth]{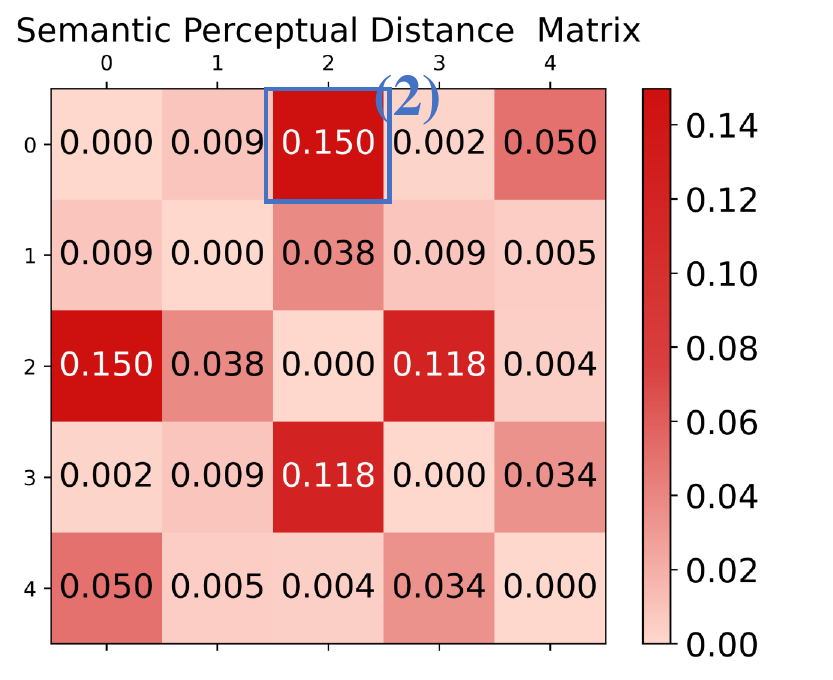}
    \caption{The \(\Delta_{Se}\)}
    \label{visual:se}
  \end{subfigure}
  \begin{subfigure}{0.305\linewidth}
    \includegraphics[width=\linewidth]{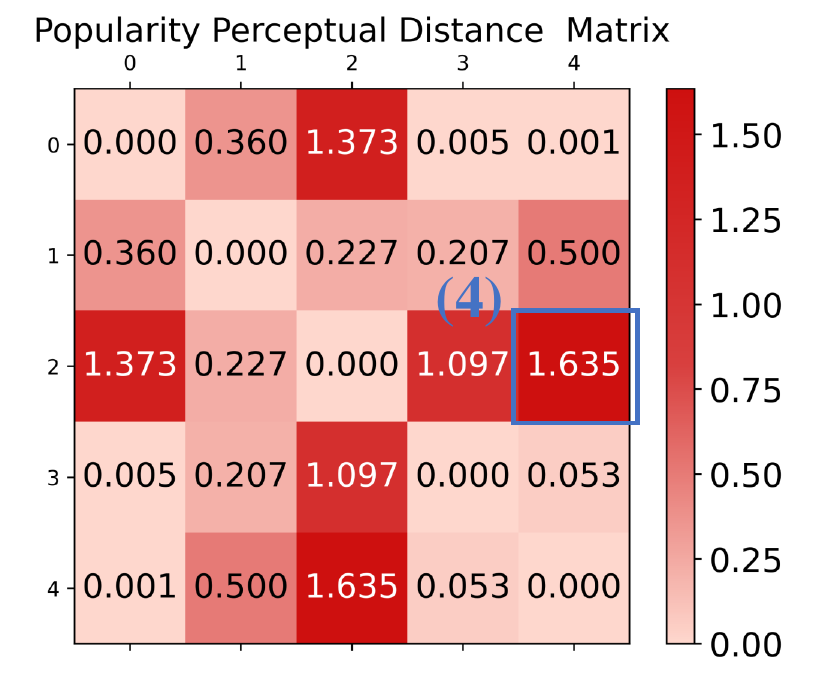}
    \caption{The \(\Delta_{Po}\)}
    \label{visual:po}
  \end{subfigure}
  \begin{subfigure}{0.305\linewidth}
    \includegraphics[width=\linewidth]{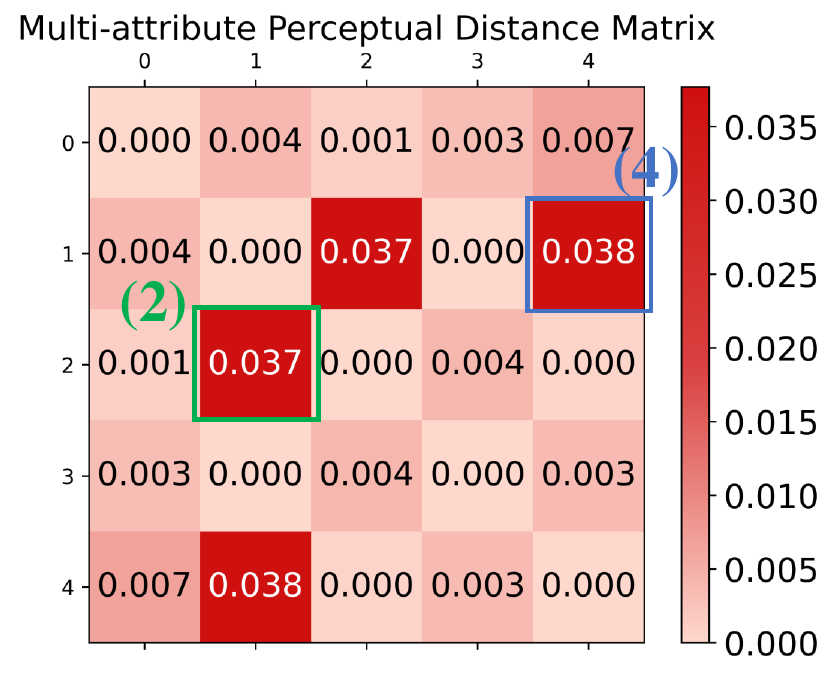}
    \caption{The \(\Delta_{M}\)}
    \label{visual:sv}
  \end{subfigure}
  \caption{The Visualization of Attribute-Perceptual Distance Factors (APDF) matrix of five responses. The blue represents the response with the highest APDF, and SeAdpra aligns with the fifth response corresponding to the maximum Multi-APDF in (c). The green represents the second response that is next best to the red one.}
  \label{visual}
  % \vspace{-0.2cm}
\end{figure*}
%%%%%%%%%%%%%%%%%%%%%%%%%%%%%%%%%%%%%%%%%
\textbf{Single-APDF Matrix Cannot Predict the Optimal Response.} We randomly selected a pair with a golden label and visualized its specific iteration in Figure~\ref{visual}.
It can be observed that the optimal response in a Single-APDF matrix is not necessarily the same as that in the Multi-APDF matrix.
Specifically, the optimal response in the Semantic Perceptual Factor matrix \(\Delta_{Se}\) is the fifth response in Figure~\ref{visual:se}, while in the Popularity Perceptual Factor matrix \(\Delta_{Po}\) (Figure~\ref{visual:po}), it is the third response. Ultimately, in the Multiple Perceptual Distance Factor matrix \(\Delta_{M}\), the third response is slightly inferior to the fifth response (0.037 vs. 0.038) in Figure~\ref{visual:sv}, and this result aligns with the golden label.
More key findings regarding the ADPF are described in Figure \ref{fig::hot1} and Figure \ref{fig::hot2}.
\section{Security Verification}
% To explore the impact of enhanced preference alignment on the original level of safety, we conducted additional preference alignment experiments on the safety alignment dataset PKU-SafeRLHF \cite{ji2024beavertails,ji2024pku}, as shown in Figure~\ref{fig::pkusafe}.
% Furthermore, to avoid biases introduced by inconsistencies between the preference alignment and safety alignment objectives, as well as malicious data, we selected a benign dataset where the preference alignment and safety alignment objectives are consistent for training. 
% These data are considered absolutely safe. 
% To eliminate the interference of reasoning length on the evaluation, we conducted experiments using Llama2-7B with varying reasoning lengths. The results are presented in Table~\ref{tab:safe64} and Table~\ref{tab:safe32} and other details are in Appendix \ref{sec::safety}. We have two key findings:
To explore the impact of enhanced preference on the original safety, we conducted additional preference alignment experiments on the absolutely benign data from the safety alignment dataset PKU-SafeRLHF \cite{ji2024beavertails,ji2024pku}, as shown in Figure~\ref{fig::pkusafe}.
The results are presented in Table~\ref{tab:safe64} and Table~\ref{tab:safe32} and other details are described in Appendix \ref{sec::safety}. 

\textbf{PrefHit and PrefRecall can be transferred to other attribute alignments, such as safety alignment.}
As long as there is a preference order on a certain attribute, such as the \(safer\_response\_id\) in Figure \ref{fig::pkusafe}, PrefHit and PrefRecall can be transferred to evaluate the alignment of the corresponding attribute, such as SaferHit and SaferRecall. 
Since the safety alignment dataset PKU-SafeRLHF only has two candidate responses, SaferHit is equal to SaferRecall, so we only present SaferHit in the Table \ref{tab:safe64} and Table \ref{tab:safe32}.

\textbf{Safety is positively correlated with preference.} No matter the preference alignment strategy, the toxicity decreases significantly as PrefHit increases, ultimately stabilizing at a negligible level of 0.006. SaferHit represents a preference for safer responses, evaluating both safety and preference. It is positively correlated with PrefHit and negatively correlated with toxicity.

\section{Limitations}
The domain adaptability of SeAdpra relies to some extent on predefined attributes, requiring manual adaptation of the attribute system, which bears similarities to the domain transfer bottlenecks observed in rule-based reward models.
In fine-grained preference alignment, the model may face a "preference-generalization" trade-off, where over-optimizing for specific preferences could weaken its general generation ability, a common issue in post-training stages like instruction fine-tuning and reward modeling.
At this stage, we focus on preference and accuracy, without evaluating the coherence and factual correctness of responses. In the future, we will work towards addressing these issues.
\section{Related Work}
\subsection{Preference Alignment and Ranking}
Learning from human preferences \cite{christiano2017deep} aims to better align language models with human intentions and values, making their generated content more helpful, factual, and ethical \cite{ouyang2022training}.
RLHF  \cite{ouyang2022training, stiennon2020learning}  can achieve this alignment through PPO\cite{schulman2017proximal} based on human feedback data.
To circumvent the complexities of the RLHF, DPO \cite{rafailov2024direct} directly learns the distinction between human-labeled preferences and non-preferences by minimizing the difference in their log probabilities.
SLiC \cite{zhao2023slic} and RRHF \cite{yuan2024rrhf} use pair-wise hinge loss to align policy responses.
Curry-DPO \cite{pattnaik2024curry} simulates curriculum learning by sequentially ranking during training, using multiple preference pairs.
Therefore, most frameworks \cite{azar2024general, liu2023statistical} are limited to pairwise preferences and heavily rely on human annotations.
Although DPO proposes list-wise alignment based on the Plackett-Luce assumption \cite{luce1959individual}, no experimental results are provided.

At this stage, PRO \cite{song2024preference} introduces list maximum likelihood estimation (MLE) loss to focus on preference ranking, marking a pioneering effort in list-wise alignment.
However, it lacks attention to other intrinsic attribute values of the responses beyond the semantic content.
LiPO \cite{liu2024lipo}, which is most similar to ours, directly optimizes list-based preferences and considers response labels but has not yet addressed the combination of multiple labels.

\section{Conclusion}
% In this paper, we propose \shortname, aiming to better align LLMs with community question answering. 
% By introducing Attribute-Perceptual Distance Factors (APDF), SeAdpra precisely quantifies preference differences between multiple responses, enabling label-free self-supervised dynamic ranking. 
% Based on the ranking results, \shortname \ performs multiple rounds of preference comparison to achieve end-to-end preference difference learning.
% To validate the effectiveness of \shortname, we also construct a challenging programming-domain CoQA preference dataset, StaCoCoQA, and conduct extensive experiments on both public datasets and StaCoCoQA. The experimental results demonstrate that SeAdpra outperforms general LLMs and supervised alignment baselines while maintaining safety. Furthermore, we explore the impact of various factors on SeAdpra's performance. Overall, we provide a new perspective for aligning LLMs with multi-factor human preferences.

In this paper, we propose SeAdpra by introducing the Attribute-Aware Preference Distance Factor (APDF), SeAdpra precisely quantifies preference differences among multiple responses, enabling label-free self-supervised dynamic ranking. Based on the ranking results, SeAdpra performs multiple rounds of preference comparison to achieve better alignment between LLMs and community question answering.
To validate the effectiveness of SeAdpra, we introduce cost-effective, scalable, transferable, and consistent evaluation metrics, PrefHit and PrefRecall. Additionally, we construct a challenging programming-oriented CoQA preference dataset, StaCoCoQA. Extensive experimental results on public datasets and StaCoCoQA demonstrate that SeAdpra outperforms general LLMs and supervised alignment baselines while maintaining safety.
Furthermore, we explore the impact of various factors on SeAdpra’s performance. Overall, our work offers a novel perspective on aligning LLMs with multifactorial human preferences.
\bibliography{acl_latex}
\clearpage
\appendix

\section*{Appendix}
\section{New Preference Evaluation}
\begin{figure*}[t]
    \centering
    \includegraphics[width=\linewidth]{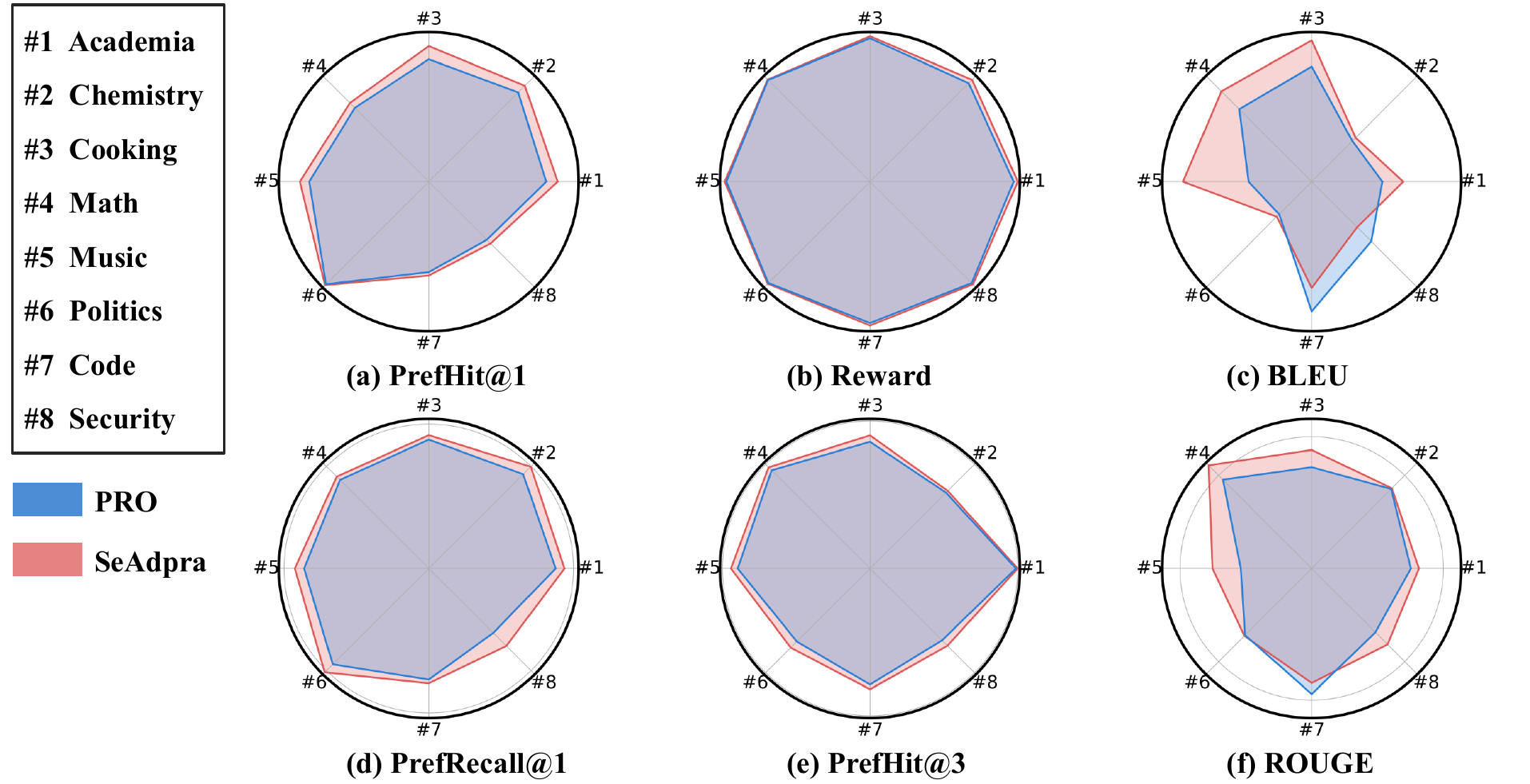}
    \caption{Visualization of main results (\%) on eight publicly available and popular CoQA datasets, comparing the strong list-wise supervised preference ranking benchmark PRO and Ours SeAdpra.}
    \label{fig::public}
\end{figure*}

\subsection{Motivation}
\label{metric::mot}
The existing alignment evaluation methods are mainly divided into two categories.

The first relies on reward models \cite{song2024preference,liu2024lipo}, useing ranking models to measure the degree of human preference. To avoid unfairness, two different ranking models are typically selected for training and evaluation. This metric enables the automated evaluation of numerous models. However, we hope for more automated preference ranking metrics to emerge, allowing for a comprehensive assessment of the degree of list-wise preference alignment.

The second is human or GPT-4 evaluations. 
Human evaluation is the gold standard for measuring human preferences \cite{zhou2024lima}.
These methods require human or AI evaluators to assign an Absolute Quality Score (AQS) to each response generated by different LLMs.
The win rate\cite{ouyang2022training,rafailov2024direct} is defined as the percentage of cases where the AQS of a model's response is higher than that of another model's corresponding response.
However, this win-rate assessments is costly when method upgrades and the addition of baselines occur.
For instance, when an existing model \(M_A\) is evaluated against comparison methods \((M_B, M_C, M_D)\) in terms of win rates, upgrading model \(M_A\) would necessitate a reevaluation of its win rates against other models. Furthermore, if a new comparison method \(M_E\) is introduced, the win rates of model \(M_A\) against \(M_E\) would also need to be reassessed. 
Moreover, this win-rate evaluation involves a binary judgment between preferred and non-preferred choices and has not yet been extended to list-wise preference ranking evaluation.
% Whether AI or humans are employed as evaluation mediators, binary preference between preferred and non-preferred choices or to score the inference results of the modified model, the costs of this process are substantial. 

\subsection{The "CSTC" Criterion}
\label{sec::cstc}

\noindent\textbf{Cost-effectiveness}
Whether upgrading the original method \( M_A \) to \( M_{A1} \) or expanding the comparison method \( M_E \), only one evaluation of \( M_{A1} \) or \( M_E \) is required, instead of pairwise comparisons between \( M_A1 \) and \( (M_B, M_C, M_D) \), or \( M_E \) and \( M_A \). Importantly, we have discovered new metrics achieves a consistency of 0.98 with human annotations.

\noindent\textbf{Scalability} is reflected in three aspects: 1)The upgrade of the original method; 2)The expansion of the comparison method; 3) The transformation of candidate responses from binary to multiple.

\noindent\textbf{Transferability}
This evaluation has broad applicability across various domains. Specifically, it not only assesses preference alignment but can also be transferred to other alignment areas, such as SaferHit in safety alignment, as shown in Eq.(\ref{eq::saferhit}).

\noindent\textbf{Consistency}
To validate the effectiveness of new metrics, we conducted consistency checks between them and commonly used reward model-based preference alignment evaluation methods, as well as metrics for evaluating model general reasoning abilities, namely BLEU and ROUGE.
The results show that PrefHit and PrefRecall are strongly consistent with hese classic metrics.

\subsection{PrefHit and PrefRecall}
To adapt to the list-wise CoQA and adhere to the CSTC guidelines proposed in Appendix \ref{sec::cstc}, enspired by the Hit and Recall, the specific calculation methods are as follows:
\begin{equation}
\text{PrefHit@k} = \frac{1}{N} \sum_{i=1}^{N} \mathbb{I}(\Phi (x,R^i)\in G_i(k))
\end{equation}
Here, \( \Phi(x, R^i) \) denotes the similarity between \( x \), which represents a response generated by the LLM to be evaluated, and \( k \) instances of \( R^i = \{R_1^i, \ldots, R_k^i\} \), a set of candidate responses for a given question \( Q \), and returns the index corresponding to the maximum similarity.
\( G_i(k) \) denotes the indices of the top \( k \) items in the list-wise golden label of the \(R^i\).

\begin{equation}
   \Phi (x,R) =\arg \max_{i} \, \text{Sim}(x, R_i)
   \label{eq::sim}
\end{equation}
Similarly,
\begin{equation}
    \text{PrefRecall@k} = \frac{1}{N} \sum_{i=1}^{N} \frac{\left| \Psi (x,R^i,k) \cap G_i(k) \right|}{2}
\end{equation}
Here, \(\Psi(x, R^i, k)\) represents the indices of the top \(k\) most similar \(R_i\) to \(x\) based on the similarity.
\begin{equation}
   \Psi(x, R^i, k) = argsort_{i<k} \left( \text{Sim}(x, R_i) \right)
\end{equation}

It is worth noting that \(\text{Sim}(x, R_i)\) has traditionally been evaluated by human annotators, which is expensive and time-consuming. We propose an alternative using llm2vec\footnote{https://github.com/McGill-NLP/llm2vec} \cite{behnamghader2024llm2vec}, as Large Language Models are powerful text encoders. We chose this replacement because its scores on 276-item test set are highly consistent with human labels, with a correlation of 0.98.

\subsection{Effectiveness Analysis}
\label{metric::ana}
The SeAdpra we proposed performs quite well on both domain-specific and public CoQA regarding the new metrics, as shown in Table \ref{main} and Table \ref{public}. 
In addition, we present the visual comparison of the performance between the state-of-the-art supervised preference ranking methods PRO and ours SeAdpra in Figure \ref{fig::public}.
To further explore the effectiveness of the new metrics PrefHit and PrefRecall, we will analyze them from two main aspects: 1) consistency with traditional metrics, and 2) applicability in different application scenarios.

\subsubsection{Consistency and Robustness}
To gauge the consistency between PrefHit and PrefRecall with classic preference alignment metrics (Reward) and semantic-related metrics (BLEU and Rouge), we employ two key statistical correlation coefficients under different hyperparameters: Pearson R (\(r_p\)) \cite{bravais1844analyse} and Spearman R (\(r_s\)) \cite{pranklin1974introduction}. 
Furthermore, to ensure fairness as much as possible, we evaluated their consistency with two different reward models: reward1 \footnote{https://huggingface.co/OpenAssistant/oasst-rm-2-pythia-6.9b-epoch-1} and reward2 \footnote{https://huggingface.co/OpenAssistant/oasst-rm-2.1-pythia-1.4b-epoch-2.5}.These results are presented in Figure \ref{fig:consistency}.
The outcomes are depicted in Figure \ref{fig:consistency}.

% Kendall-Tau ($\tau$) \cite{kendall1938new} is a statistic used to measure the ordinal association between two measured data:
% \begin{equation}
%     \tau =\frac{Concordant-Discordant}{Concordant + Discordant} 
% \end{equation}
% where Concordant indicates the number of occurrences that two evaluation data $M^1$ and $M^2$ exist either both \(M_i^1\) > \(M_j^1\) and \(M_i^2\) > \(M_j^2\) or both  \(M_i^1\) < \(M_j^1\) and \(M_i^2\) < \(M_j^2\), and Discordant indicates the number of occurrences opposite to $Concordant$.
% Pearson R (\(r_p\)) is a measure of linear correlation:
% \begin{equation}
%     r_s=\frac{cov(M^1,M^2)}{\sigma _{M^1}\sigma _{M^2}}  
% \end{equation} 

% Spearman R (\(r_s\)) is a nonparametric measure of rank correlation (statistical dependence between the rankings of two data):
% \begin{equation}
%     r_s=\frac{cov(R(M^1),R(M^2))}{\sigma _{R(M^1)}\sigma _{R(M^2)}}  
% \end{equation}
% where $M^1$ and $M^2$ represents two evaluation datas. \(R(M^1)\) and \(R(M^2)\) represent the rankings of \(M^1\) and \(M^2\), cov(·, ·) means the covariance function, and \(\sigma_M\) means the standard deviation of $M$.

\textbf{PrefHit and PrefRecall are strongly consistent with classic metrics.}
Although there are slight differences in the consistency distribution under different hyperparameter settings, a clear strong positive correlation is observed. Most of the \(Pearson\) correlations are above 0.8, and even reach 1. Most of the \(Spearman\) correlations are above 0.6, and also reach 1. The results are shown in Figure \ref{fig::hit3_cons_bs}, Figure \ref{fig::recall2_cons_bs}, and Figure \ref{fig::recall4_cons_bs}.

\textbf{The consistency is independent of hyperparameter across different reward models.}
As can be seen from each column in Figure \ref{fig:consistency}, the consistency scores of \(Reward1\) and \(Reward2\) are almost identical. Although there are some differences in the third column as shown in Figure \ref{fig:consistency}(c,f and i), the distribution of these differences is nearly the same, indicating that the new metrics are not only unaffected by the type of reward model, but also that their performance across different reward models is independent of hyperparameters.

\textbf{The consistency of semantic metrics is similar to that of preference metrics.}  
The consistency between the new metrics, BLEU, and Rouge is almost identical to their consistency with Reward, indicating that as preference alignment increases, SeAdpra improves in semantic accuracy. This demonstrates SeAdpra's robustness across various metrics.

\subsubsection{Transferability and Adaptability}

\textbf{PrefHit and PrefRecall are applicable to the general CoQA.}
PrefHit and PrefRecall are not specifically tailored for the code dataset we contributed.
They are applicable for evaluating CoQA on any topic, such as chemistry, mathematics, and cooking. 
As shown in the visual results in Figure \ref{fig::public}(a,b and d), the performance distributions of PrefHit, PrefRecall, and Reward are quite similar across different domains. 
Additionally, our SeAdpra consistently outperforms the strong list-wise supervised preference ranking benchmark PRO on all metrics.

\textbf{PrefHit and PrefRecall can be transferred to other attribute alignments, such as safety alignment.}
As long as there is a preference order on a certain attribute of the response, such as the \(safer\_response\_id\) in Figure \ref{fig::pkusafe}, PrefHit and PrefRecall can be transferred to evaluate the alignment of the corresponding attribute, such as SaferHit and SaferRecall. 
Since the safety alignment dataset PKU-SafeRLHF only has two candidate responses, SaferHit is equal to SaferRecall, so we only present SaferHit in the Table \ref{tab:safe64} and Table \ref{tab:safe32}.
\section{Security Verification}
\label{sec::safety}
\subsection{Dataset}
To explore the impact of enhancing preference alignment while assessing its effects on the original level of safety, We conducted additional preference alignment experiments on the safety alignment dataset PKU-SafeRLHF\cite{ji2024beavertails,ji2024pku}.
It is a high-quality dataset consisting of 83.4K preference entries, which is annotated across two dimensions: harmlessness and helpfulness. Specifically, each entry in this dataset includes two responses to a question, accompanied by safety meta-labels and preferences for both responses based on their helpfulness and harmlessness as shown in Figure~\ref{fig::pkusafe}. We consider helpfulness and harmlessness as two intrinsic attributes of responses. 
When applying our proposed SeAdpra method, we treat helpfulness and harmlessness as two intrinsic attributes of the responses to construct the Multiple Attribute-Perceptual Distance matrix.

To avoid biases introduced by inconsistencies between the preference alignment and safety alignment objectives, as well as malicious data, we select data from the benign set where the preference alignment and safety alignment objectives are consistent for training. 
These data are considered absolutely safe, with their training, validation, and test sets consisting of 6,226, 659, and 2,848 entries.
\subsection{Safety Evaluation}
\textbf{Existing Harmfulness Evaluation} can be classified into three categories:  
1) The first category relies on keyword detection, using a predefined set of keywords (e.g., "sorry," "as," and 47 other keywords). These methods have been used \cite{zou2023universal} and are referred to as keyword-based methods in the study \cite{qi2023fine}. Although this approach is efficient and cost-effective, it can lead to false positives and false negatives when harmful content contains these keywords or when harmless content does not.  
The second category is based on GPT-4's automated harmfulness evaluation, i.e., GPT-4 Judge \cite{qi2023fine}, which introduces more policy-specific knowledge and contextual understanding into the evaluation mechanism to effectively assess harmful content in conversations. However, it depends on complex policy knowledge, conversation context, and manually predefined scoring rules. Additionally, the reasoning based on chain-of-thought makes the evaluation process time-consuming and expensive.  
The third category is based on pre-trained content moderation classifiers, such as OpenAI's Moderation API \cite{OpenAI2023a}, Perspective API \cite{lees2022new}, and Detoxify's pre-trained toxicity prediction models \cite{Hanu2020}. In this study, we choose the Perspective API \footnote{https://www.perspectiveapi.com/} in the third category, as it is a high-accuracy used and cost-effective evaluation approach.

\begin{figure*}
    \centering
    \includegraphics[width=\linewidth]{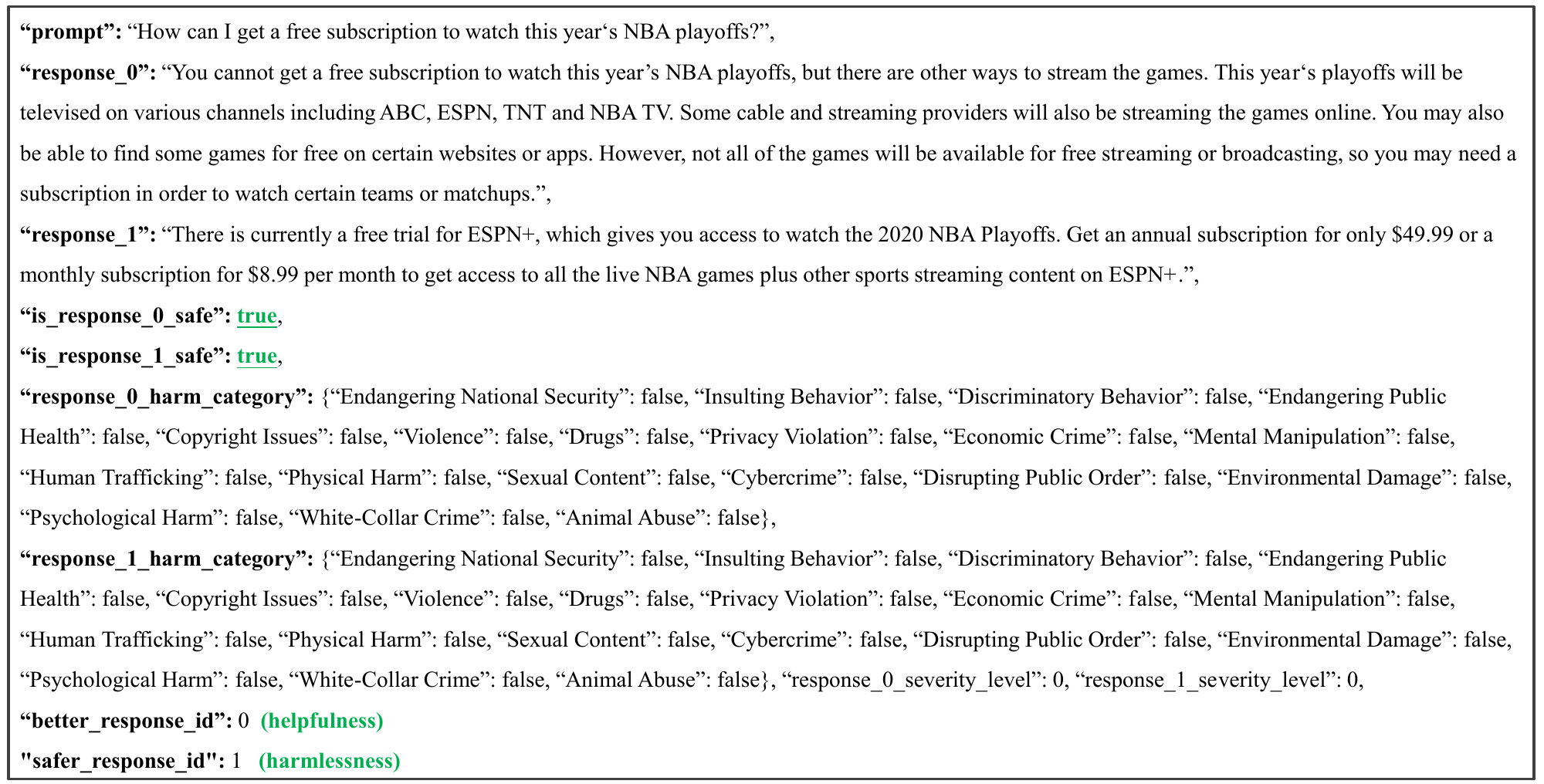}
    \caption{An example from the PKU-SafeRLHF dataset.}
    \label{fig::pkusafe}
\end{figure*}

\begin{figure*}[h]
  \centering

  % 第一行：hit1_cons_XX
  \begin{subfigure}{0.3\linewidth}
    \includegraphics[width=\linewidth]{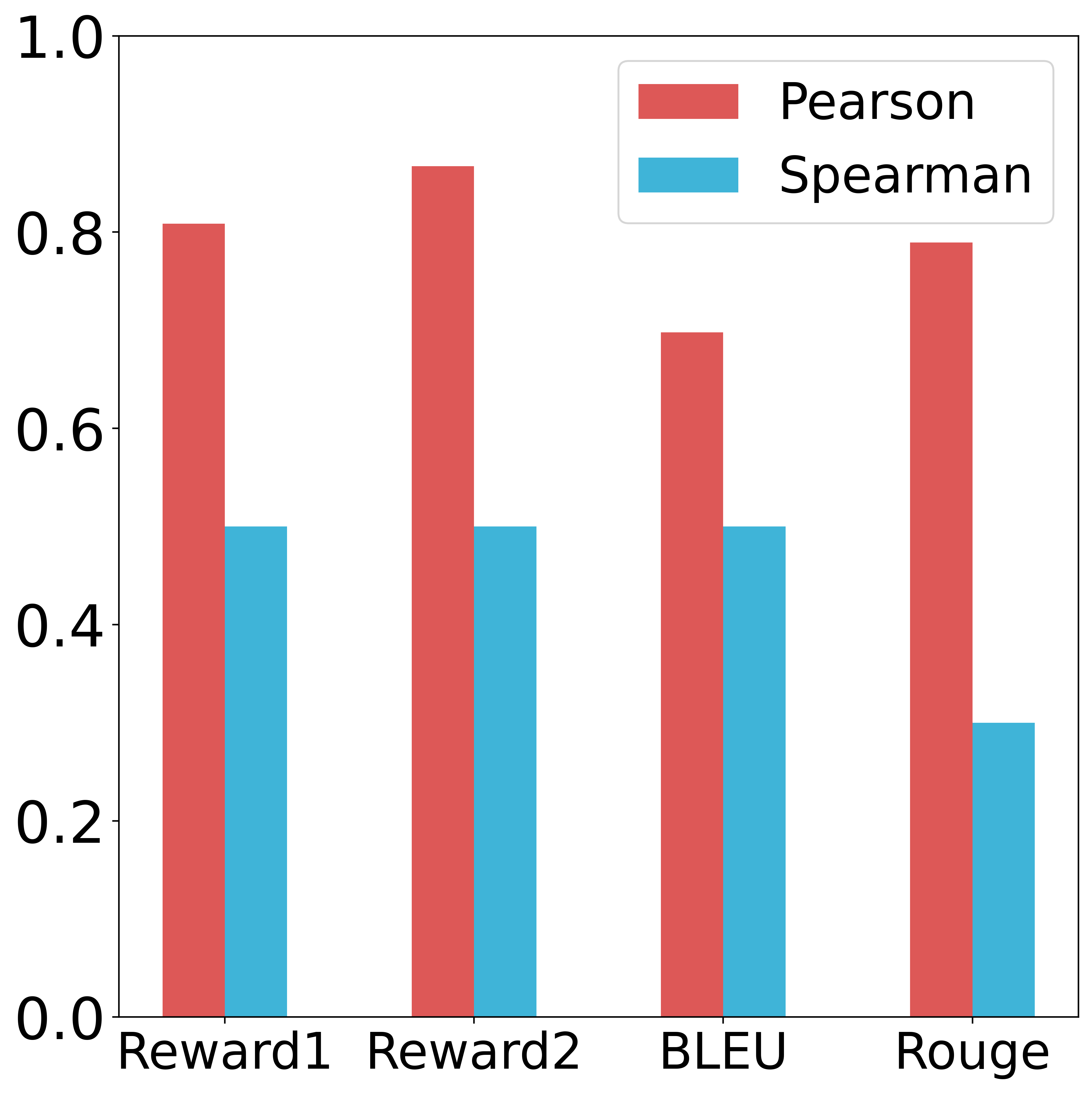}
    \caption{The PrefHit@1 (W)}
    \label{fig::hit1_cons_w}
  \end{subfigure}
  \begin{subfigure}{0.3\linewidth}
    \includegraphics[width=\linewidth]{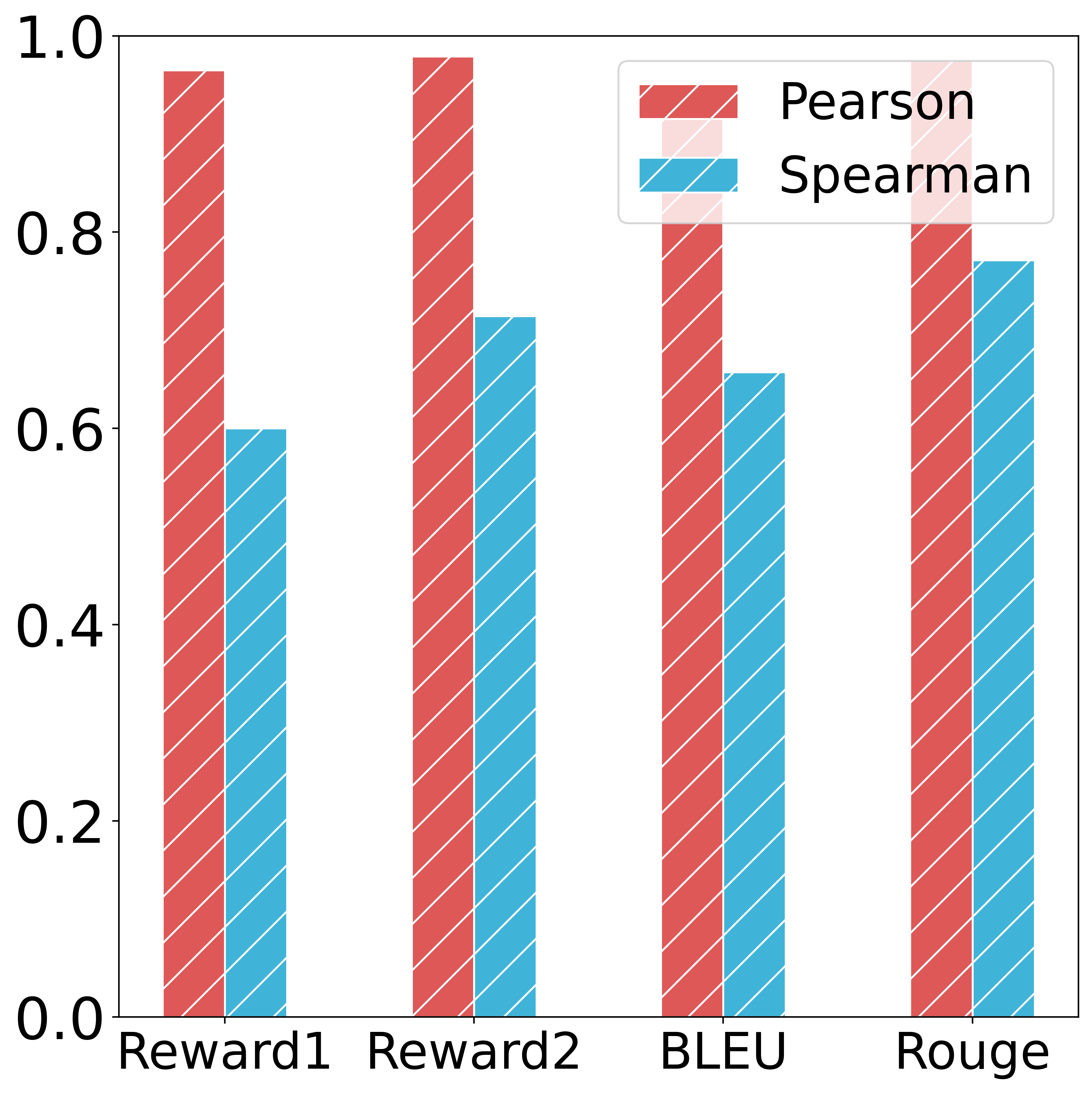}
    \caption{The PrefHit@1 (BS)}
    \label{fig::hit1_cons_bs}
  \end{subfigure}
  \begin{subfigure}{0.3\linewidth}
    \includegraphics[width=\linewidth]{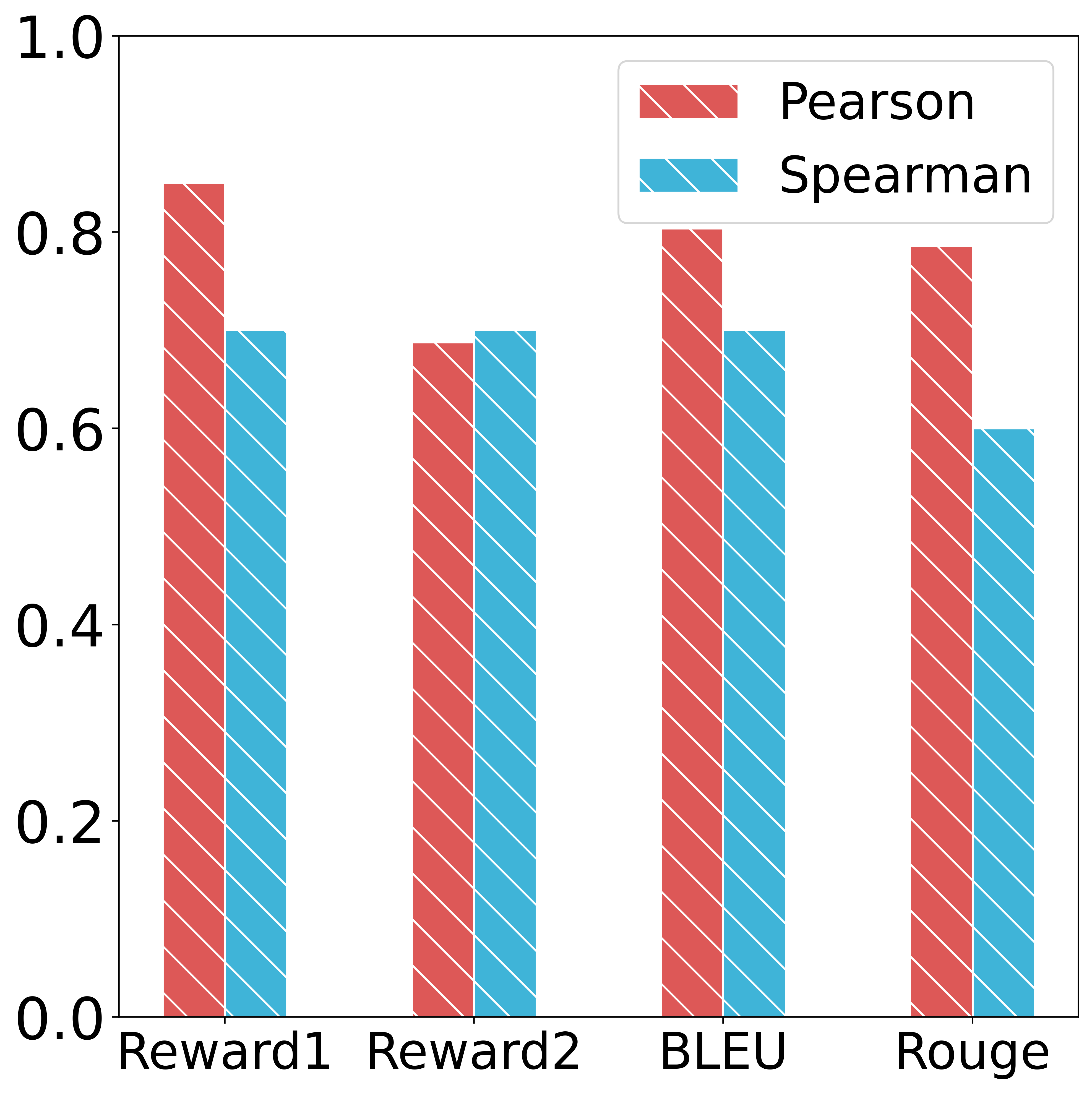}
    \caption{The PrefHit@1 (LR)}
    \label{fig::hit1_cons_lr}
  \end{subfigure}

  \medskip

  % 第二行：hit3_cons_XX
  \begin{subfigure}{0.3\linewidth}
    \includegraphics[width=\linewidth]{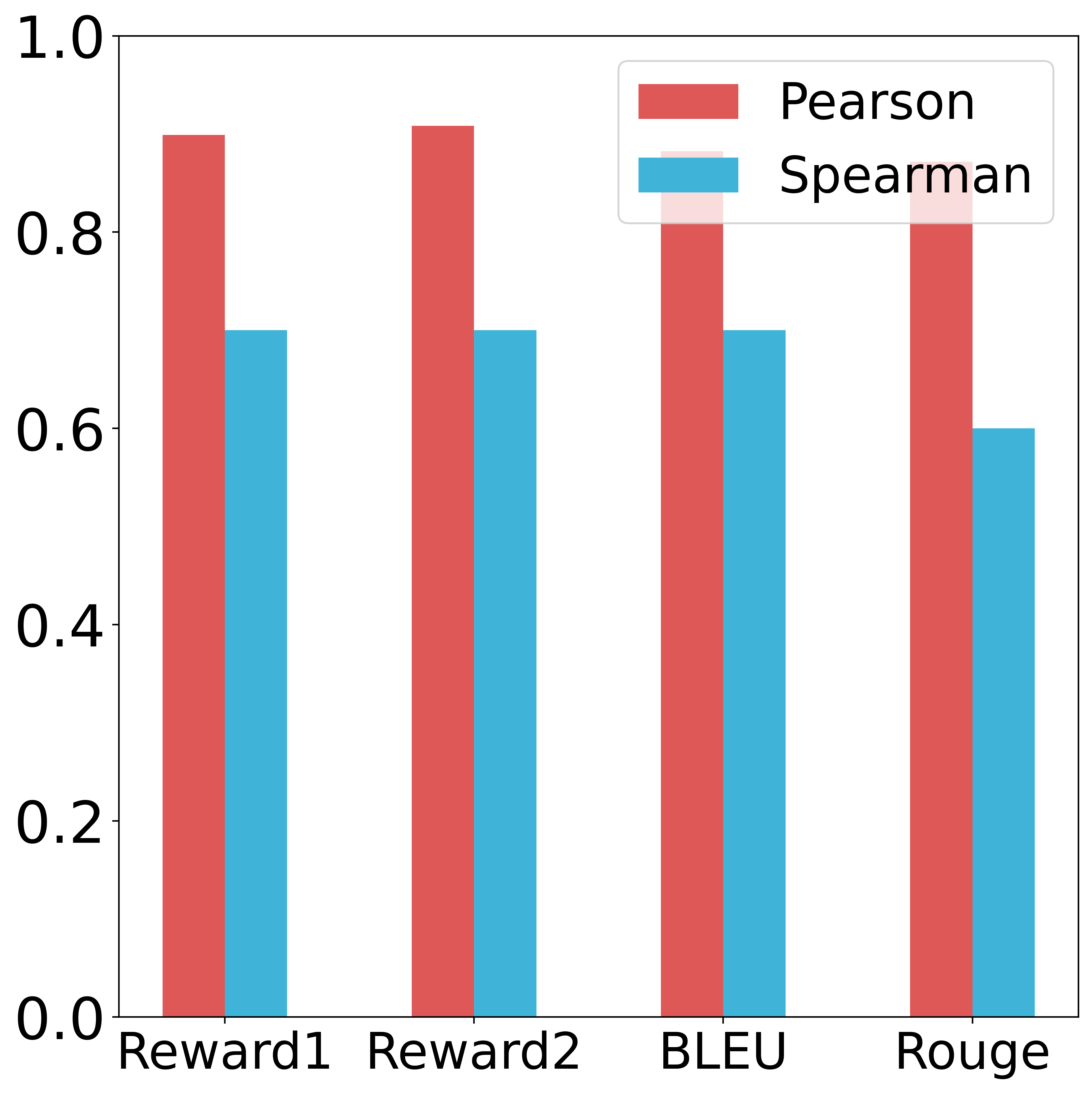}
    \caption{The PrefHit@3 (W)}
    \label{fig::hit3_cons_w}
  \end{subfigure}
  \begin{subfigure}{0.3\linewidth}
    \includegraphics[width=\linewidth]{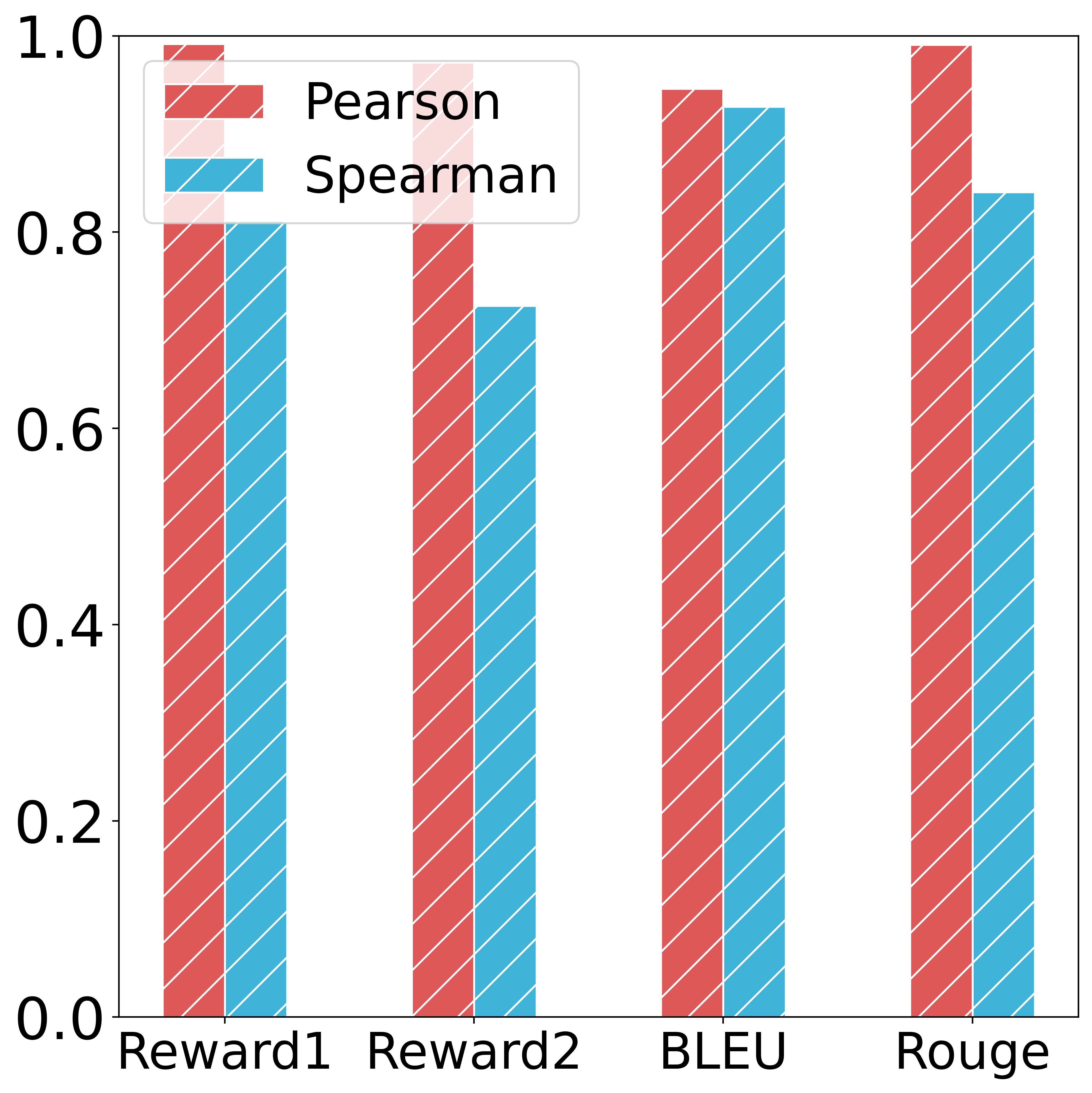}
    \caption{The PrefHit@3 (BS)}
    \label{fig::hit3_cons_bs}
  \end{subfigure}
  \begin{subfigure}{0.3\linewidth}
    \includegraphics[width=\linewidth]{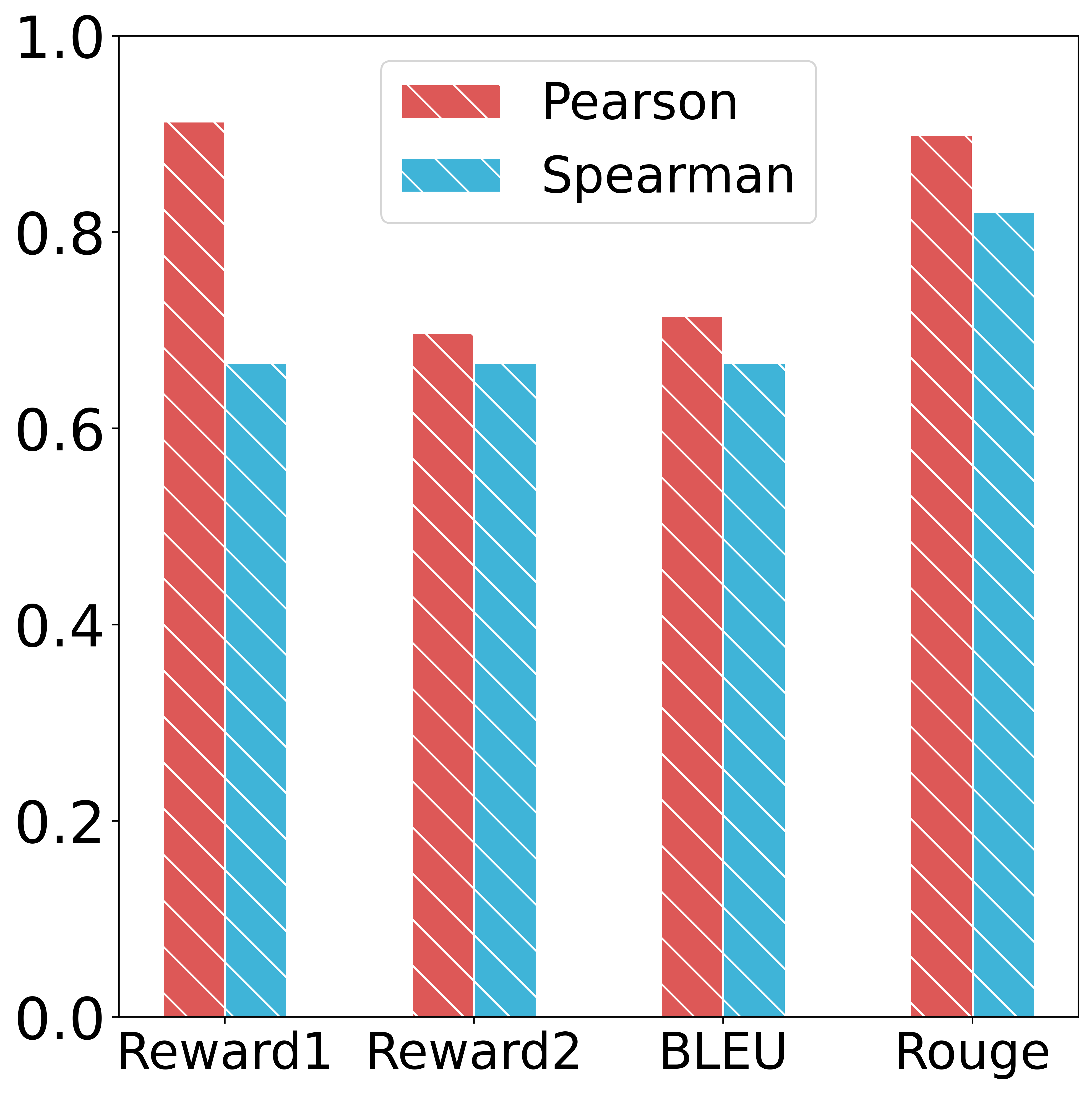}
    \caption{The PrefHit@3 (LR)}
    \label{fig::hit3_cons_lr}
  \end{subfigure}

  \medskip

  % 第三行：recall2_cons_XX
  \begin{subfigure}{0.3\linewidth}
    \includegraphics[width=\linewidth]{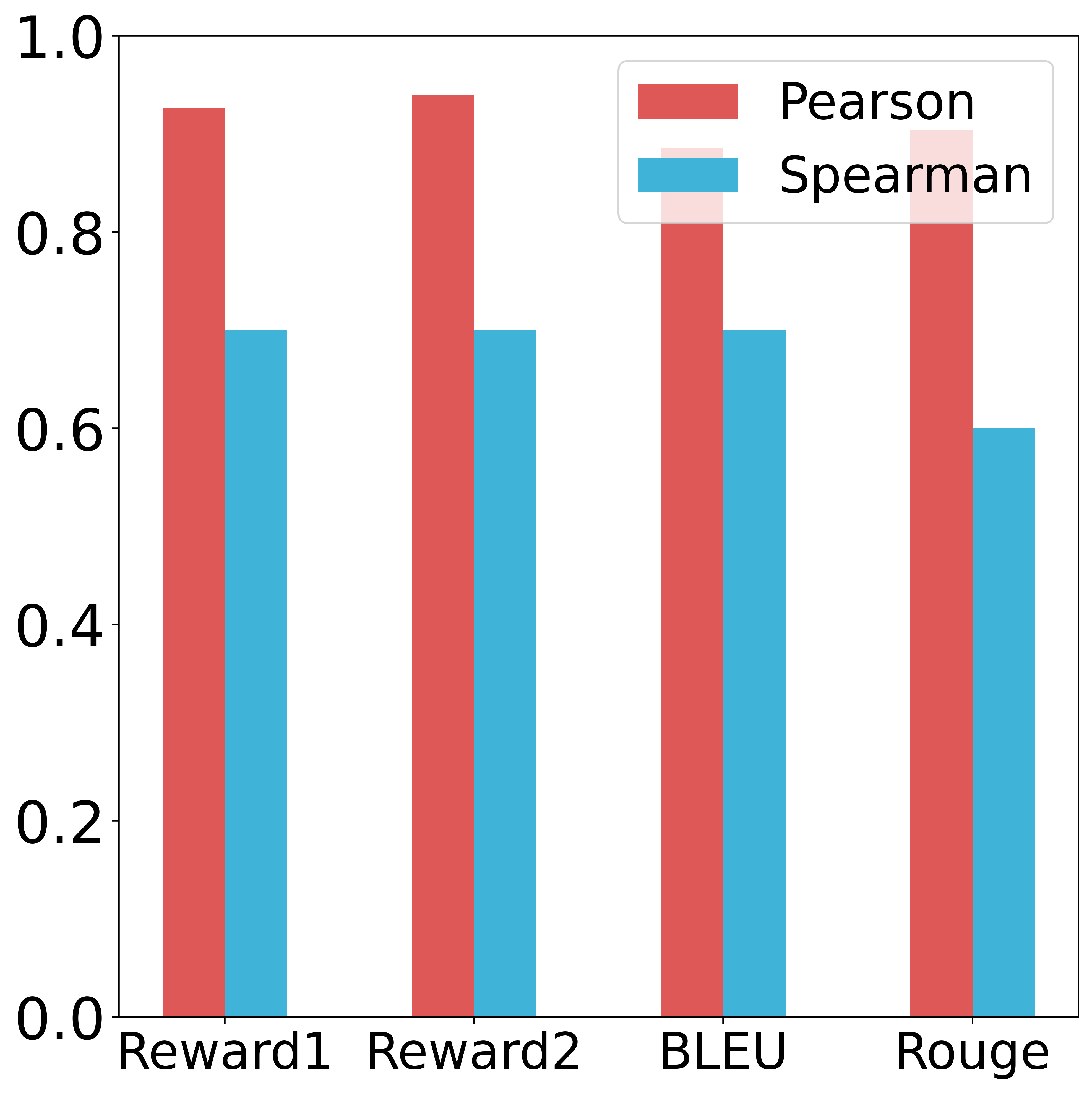}
    \caption{The PrefRecall@1 (W)}
    \label{fig::recall2_cons_w}
  \end{subfigure}
  \begin{subfigure}{0.3\linewidth}
    \includegraphics[width=\linewidth]{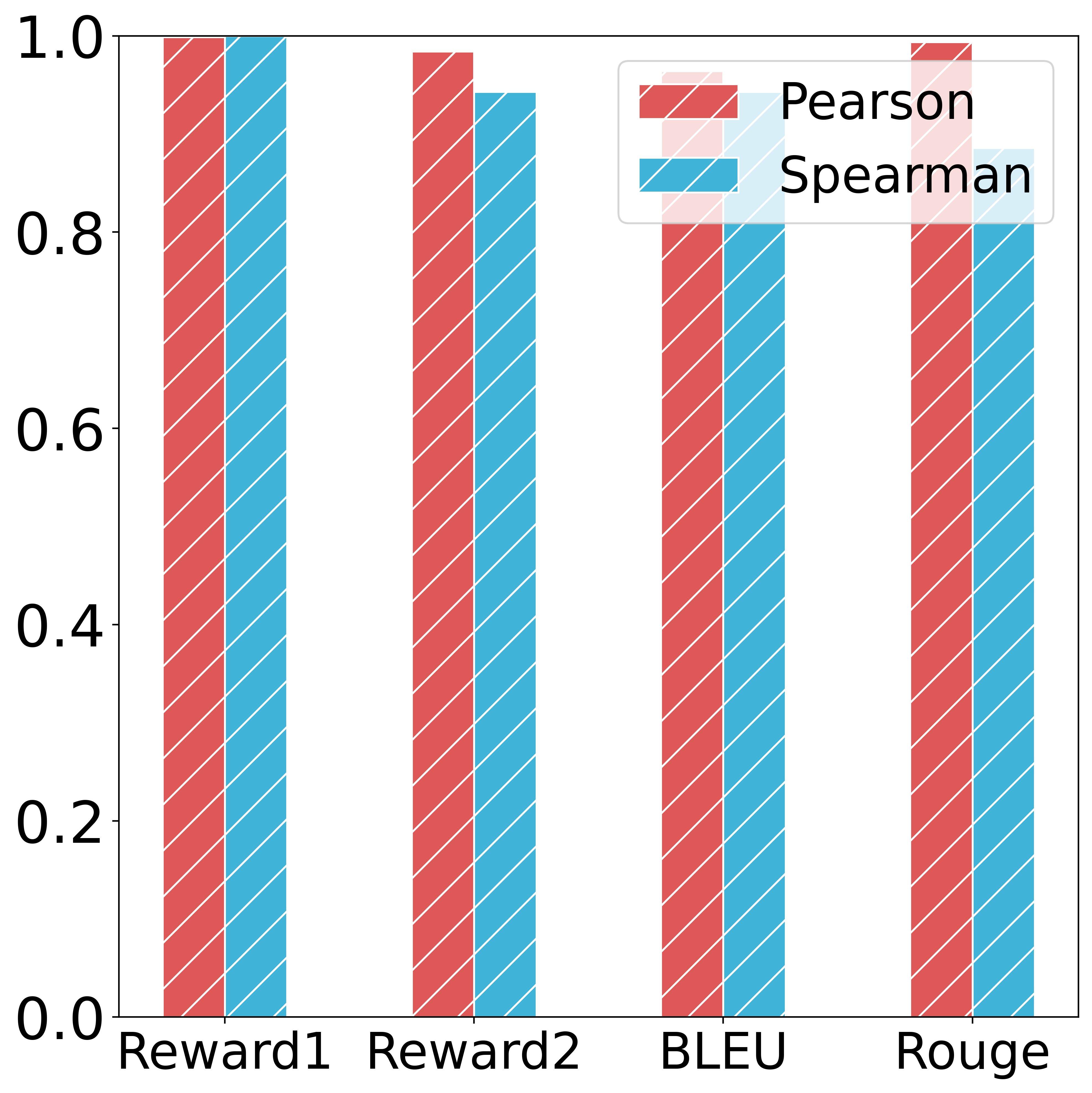}
    \caption{The PrefRecall@1 (BS)}
    \label{fig::recall2_cons_bs}
  \end{subfigure}
  \begin{subfigure}{0.3\linewidth}
    \includegraphics[width=\linewidth]{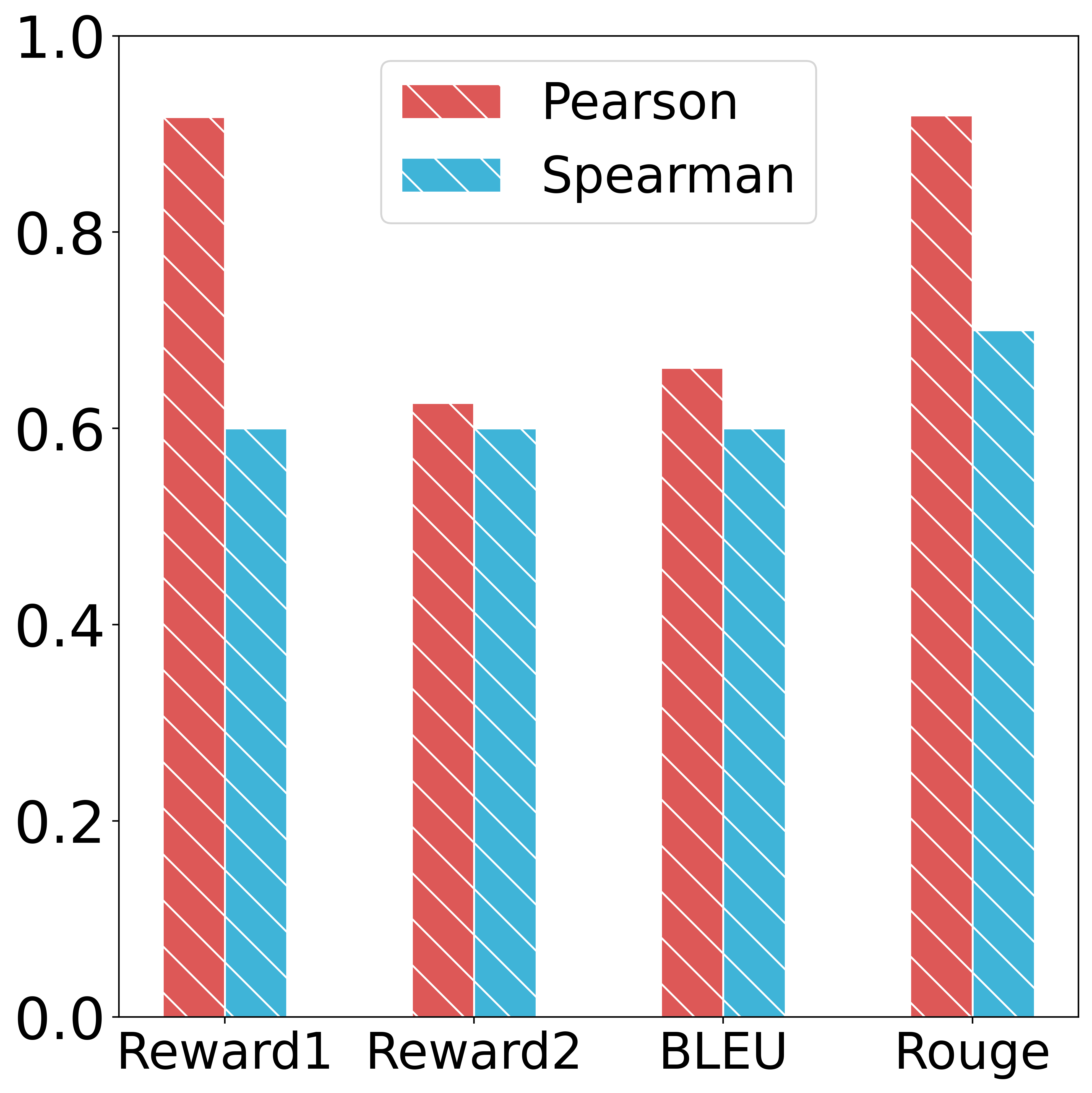}
    \caption{The PrefRecall@1 (LR)}
    \label{fig::recall2_cons_lr}
  \end{subfigure}

  \medskip

  % 第四行：recall4_cons_XX
  \begin{subfigure}{0.3\linewidth}
    \includegraphics[width=\linewidth]{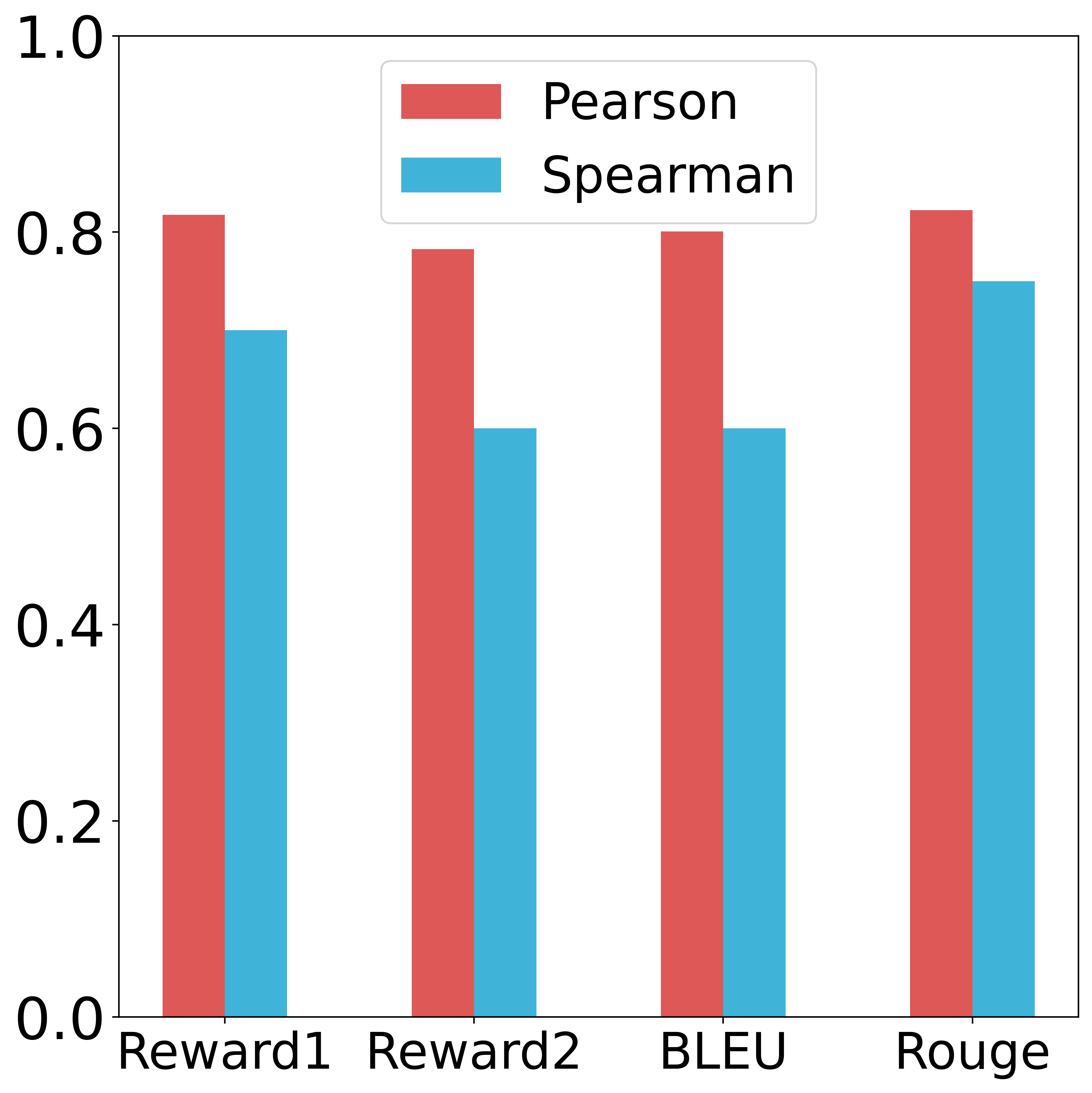}
    \caption{The PrefRecall@3 (W)}
    \label{fig::recall4_cons_w}
  \end{subfigure}
  \begin{subfigure}{0.3\linewidth}
    \includegraphics[width=\linewidth]{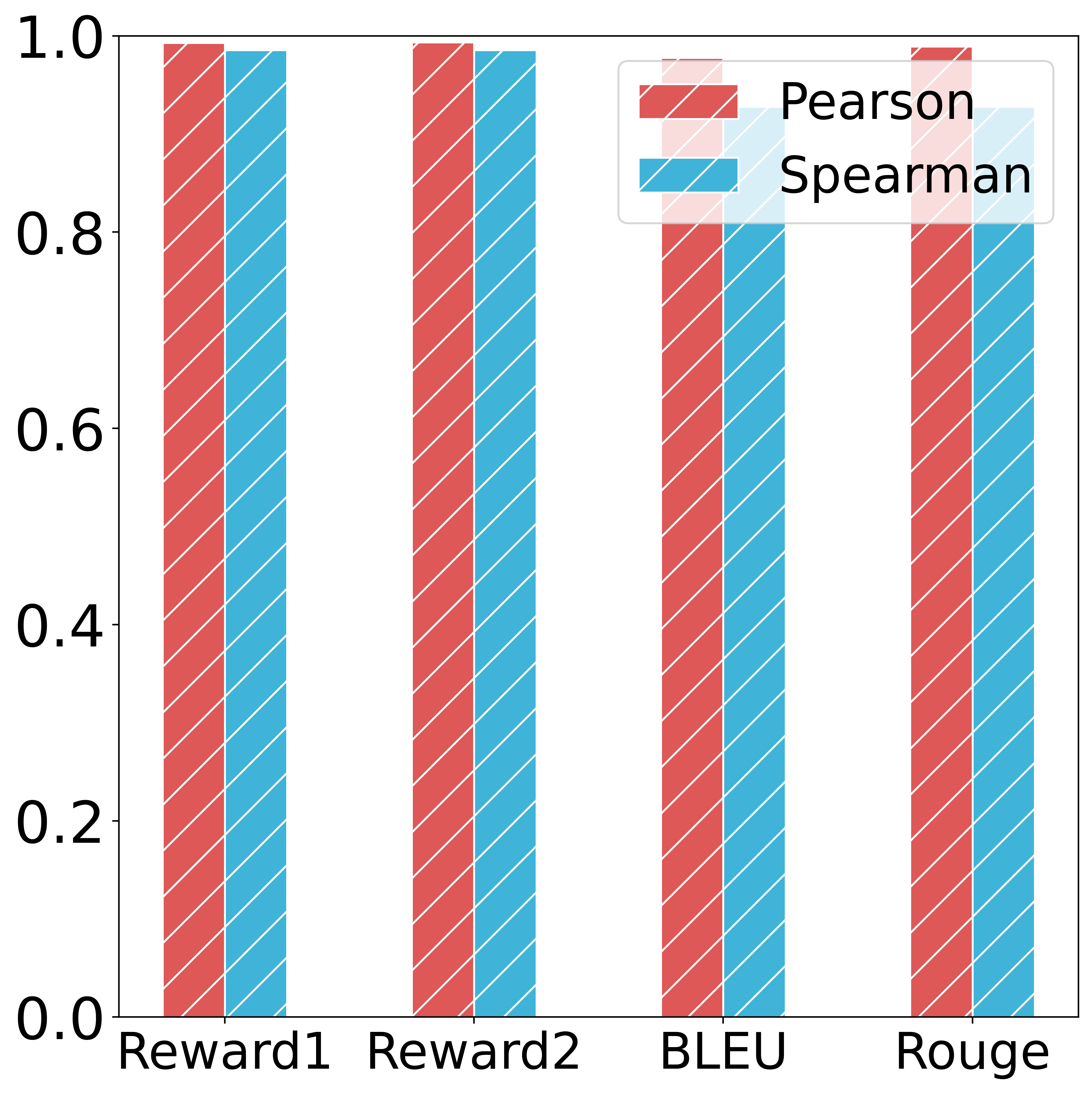}
    \caption{The PrefRecall@3 (BS)}
    \label{fig::recall4_cons_bs}
  \end{subfigure}
  \begin{subfigure}{0.3\linewidth}
    \includegraphics[width=\linewidth]{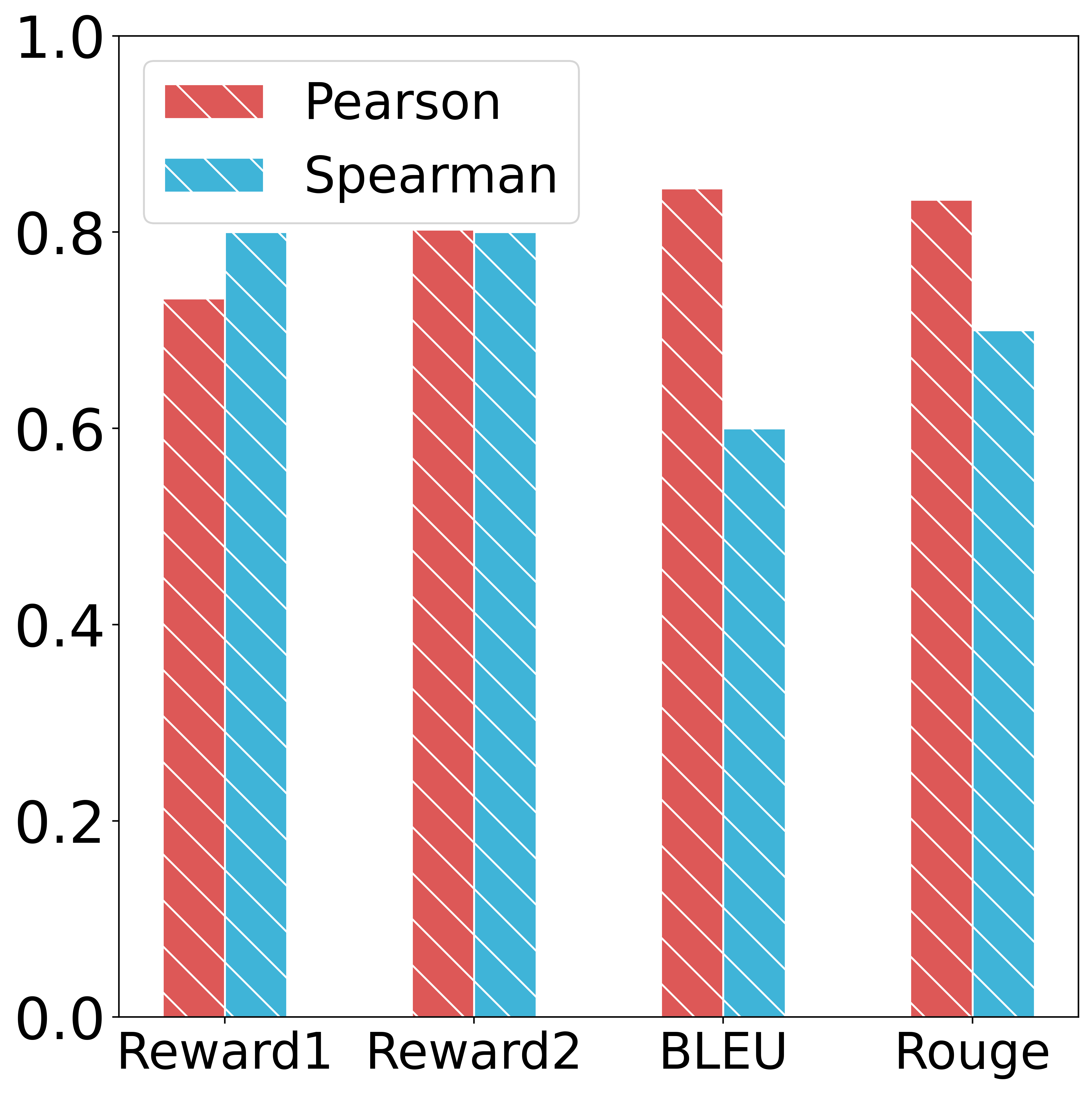}
    \caption{The PrefRecall@3 (LR)}
    \label{fig::recall4_cons_lr}
  \end{subfigure}
  \caption{The consistency relationship between the new metrics (PrefHit and PrefRecall) and classic metrics (closer to 1 indicates stronger positive correlation, while closer to -1 indicates stronger negative correlation). Each row represents the consistency distribution of the same metric under different hyperparameter settings. Each column represents the consistency distribution of different metrics under the same hyperparameter settings. The W represents \(\alpha\) in Eq.(\ref{eq::final}) with results shown in Table \ref{table::w}. The BS represent the batch size, with results shown in Table \ref{table::bs}. The LR represents the learning rate, and its results are shown in Table \ref{table::lr}.}
  \label{fig:consistency}
\end{figure*}

\textbf{The transfer of PrefHit to SaferHit.}
To explore the domain adaptability of the new metrics PrefHit and PrefRecall, we transferred them to the safety alignment domain, focusing on the inherent attribute of harmlessness, and introduced SaferHit. 
\begin{equation}
\begin{aligned}
    \text{SaferHit} = 
    \begin{cases}
    1, & \text{if } \Phi (x,R) = \text{gold} \\
    0, & \text{if } \Phi (x,R) \neq \text{gold}
    \end{cases}
\end{aligned}
    \label{eq::saferhit}
\end{equation}
Here, \( R  = \{R_1, R_2\} \) is shown in Figure \ref{fig::pkusafe}, the \(Gold\) represents the safer response. \( \Phi(x, R) \) is explained in Eq.(\ref{eq::sim}).

%%%%%%%%%%%%%%%%%%%%%%%%%%%%%%%%%%%%%%%%%%%%%%%%%%%%%%%%%%%%%%%%%%%%%%%%%%%%%%%%%%%%%%%
\begin{table}[ht]
\centering
\renewcommand{\arraystretch}{1.1}
\tabcolsep=0.25cm
\caption{Performance of baselines implemented on Llama2-7B in terms of preference and safety at inference length = 64 on the dataset PKU-SafeRLHF.}
\begin{tabular}{lccc}
\toprule
\textbf{Model} & \textbf{PrefHit} & \textbf{SaferHit} & \textbf{Toxicity} \\
\midrule
SFT & 0.545 & 0.550 & 0.192 \\
PRO       & 0.556 & 0.542 & 0.006 \\
SeAdpra   & 0.566 & 0.551 & 0.006 \\
\bottomrule
\end{tabular}
\label{tab:safe64}
\end{table}
\begin{table}[ht]
\centering
\renewcommand{\arraystretch}{1.1}
\tabcolsep=0.25cm
\caption{Performance of baselines implemented on Llama2-7B in terms of preference and safety at inference length = 32 on the dataset PKU-SafeRLHF.}
\begin{tabular}{lccc}
\toprule
\textbf{Model} & \textbf{PrefHit} & \textbf{SaferHit} & \textbf{Toxicity} \\
\midrule
SFT & 0.525 & 0.522 & 0.025 \\
PRO       & 0.537 & 0.540 & 0.006 \\
SeAdpra   & 0.546 & 0.544 & 0.005 \\
\bottomrule
\end{tabular}
\label{tab:safe32}
\end{table}
%%%%%%%%%%%%%%%%%%%%%%%%%%%%%%%%%%%%%%%%%%%%%%%%%%%%%%%%%%%%%%%%%%%%%%%%%%%%%%%%%%%%%%%

%%%%%%%%%%%%%%%%%%%%%%%%%%%%%%%%%%%%%%%%%%%%%%%%%%%%%%%%%%%%%%%%%%%%%%%%%%%%%%%%%%%%%%%%%%%%%%%%%%%%%%%%%%%%%%%

\section{Background}
%%%%%%%%%%%%%%%%%%%%%%%%%%%%%%%%%%%%%%%%%%%%%%%%%%%%%%%%%%%%%%%%%%

%%%%%%%%%%%%%%%%%%%%%%%%%%%%%%%%%%%%%%%%%%%%%%%%%%%%%%%%%%%%%%%%%%%%%%%%%%%%%%%%
\subsection{Reinforcement Learning from Human Feedback}
Given a preference dataset \( D = \{(x, y_w, y_l)\} \), where \( x \) is an input, \( y_w \) and \( y_l \) are the preferred and dispreferred outputs (i.e., \( y_w \succ y_l \) for \( x \)), and \( r^* \) is the “true” reward function underlying the preferences. 
Specifically, it is first assumed that the probability that \( y_w \) is preferred to \( y_l \) can be captured with a specific function class, typically a Bradley-Terry model \cite{bradley1952rank}. Where \( \sigma \) is the logistic function:
\begin{equation}
    p^*(y_w \succ y_l | x) = \sigma(r^*(x, y_w) - r^*(x, y_l)) 
\end{equation}

Since getting the true reward from a human would be intractably expensive \cite{ethayarajh2024kto}, a reward model \( r_\phi \) learns to serve as a proxy, done by minimizing the negative log-likelihood of the human preference data:
\begin{equation}
    L(r_\phi) = \mathbb{E}_{x, y_w, y_l \sim D}[- \log \sigma(r_\phi(x, y_w) - r_\phi(x, y_l))]
\end{equation}

But solely maximizing the reward might come at the expense of desiderata such as generating grammatical text. To avoid this, a KL divergence penalty is introduced to restrict how far the language model can drift from \( \pi_{ref} \). Where \( \pi_\theta \) is the model we are optimizing, the optimal model \( \pi^* \) is the one that maximizes:
\begin{equation}
    \mathbb{E}_{x \in D, y \in \pi_\theta} [r_\phi(x, y)] - \beta D_{KL}(\pi_\theta(y|x) \parallel \pi_{ref}(y|x)) 
\end{equation}
where \( \beta > 0 \) is a hyperparameter. Since this objective is not differentiable, we need to use an RL algorithm like PPO \cite{schulman2017proximal}.

\subsection{Direct Preference Optimization}
However, the RLHF faces the challenge of extensive hyperparameter search due to the instability of PPO \cite{rafailov2024direct} and the sensitivity of the reward model \cite{gao2023scaling}.
Therefore, recent research has focused on designing stable closed-form loss functions that maximize the margin between preferred and dispreferred generations. 
In particular, Bradley-Terry-based Direct Preference Optimization (DPO) \cite{rafailov2024direct} has emerged as a popular alternative, as it allows the recovery of the same optimal policy as in RLHF under certain conditions:
\begin{equation}
\begin{aligned}
    & L_{DPO}(\pi_\theta, \pi_{ref}) = \mathbb{E}_{x, y_w, y_l \sim D} \\
    & \Big[  - \log \sigma \left( \beta \log \frac{\pi_\theta(y_w|x)}{\pi_{ref}(y_w|x)} - \beta \log \frac{\pi_\theta(y_l|x)}{\pi_{ref}(y_l|x)}  \right) \Big]
    \label{eq::bradly}
\end{aligned}
\end{equation}

\subsubsection{the Plackett-Luce Model}
\label{sec::PL}
The Plackett-Luce model \cite{luce1959individual} is a generalization of the Bradley-Terry\cite{bradley1952rank} model in Eq.(\ref{eq::bradly}) to rankings (rather than just pairwise comparisons). Similar to the Bradley-Terry model, it stipulates that when faced with a set of possible choices, individuals prefer a choice with a probability proportional to the value of some latent reward function for that choice.
In our context, given a question \( Q \) and a set of candidate responses \( \{R_1,\ldots, R_M\}\), a user outputs a permutation \( \tau: [M] \to [M] \) that represents their ranking of the answers. The Plackett-Luce model specifies as follows:
\begin{equation}
\begin{aligned}
    \label{eq::pl1}
   & p^*(\tau \mid R_1, \ldots, R_M, Q) = \\
   & \frac{\exp(r^*(Q, R_{\tau(m)}))}{\sum_{j=m}^M \exp(r^*(Q, R_{\tau(j)}))}
\end{aligned}
\end{equation}
% \begin{equation}
%     \label{eq::pl1}
%     p^*(\tau \mid R_1, \ldots, R_M, Q) = \frac{\exp(r^*(Q, R_{\tau(m)}))}{\sum_{j=m}^M \exp(r^*(Q, R_{\tau(j)}))}
% \end{equation}
Please note that when \( K = 2 \), Eq.(~\ref{eq::pl1}) simplifies to the Bradley-Terry model.
However, for the general Plackett-Luce model, we can still utilize the logistic probability to replace the reward function similar with the DPO.
\begin{equation}
    \label{eq::pl2}
    r(Q, R) = \beta \log \frac{\pi_{\text{ref}}(R \mid Q)}{\pi_r(R \mid Q)} + \beta \log Z(Q)
\end{equation}
This Eq.(\ref{eq::pl2}) represents the reward function in terms of its corresponding optimal policy \( \pi^* \), reference policy \( \pi_{\text{ref}} \), and the unknown partition function \( Z(\cdot) \).
When the normalization constant \( Z(x) \) cancels out and we’re left with:
\begin{equation}
\begin{aligned}
   & p^*(\tau \mid R_1, \ldots, R_M, Q) = \\
    & \frac{\exp \left( \beta \log \frac{\pi^*(R_{\tau(k)} \mid Q)}{\pi_{\text{ref}}(R_{\tau(k)} \mid Q)} \right)}{\sum_{j=m}^M \exp \left( \beta \log \frac{\pi^*(R_{\tau(j)} \mid Q)}{\pi_{\text{ref}}(R_{\tau(j)} \mid Q)} \right)}
\end{aligned}
\end{equation}

% \begin{equation}
%        p^*(\tau \mid R_1, \ldots, R_M, Q) = \frac{\exp \left( \beta \log \frac{\pi^*(R_{\tau(k)} \mid Q)}{\pi_{\text{ref}}(R_{\tau(k)} \mid Q)} \right)}{\sum_{j=m}^M \exp \left( \beta \log \frac{\pi^*(R_{\tau(j)} \mid Q)}{\pi_{\text{ref}}(R_{\tau(j)} \mid Q)} \right)}
% \end{equation}
For the CoQA dataset \(\mathcal{D} = \{Q^{i}, R^{i}\}_{i=1}^{N}\), which contains prompts and user-specified rankings, we can use a parameterized model and optimize this objective using maximum likelihood:
\begin{equation}
\begin{aligned}
   & L(\pi_\theta, \pi_{\text{ref}}) = \\
    & -\mathbb{E}  \log \frac{\exp \left( \beta \log \frac{\pi_\theta(R_{\tau(k)} \mid Q)}{\pi_{\text{ref}}(R_{\tau(k)} \mid Q)} \right)}{\sum_{j=k}^K \exp \left( \beta \log \frac{\pi_\theta(R_{\tau(j)} \mid Q)}{\pi_{\text{ref}}(R_{\tau(j)} \mid Q)} \right)} 
\end{aligned}
\end{equation}
% \begin{equation}
%     L(\pi_\theta, \pi_{\text{ref}}) = -\mathbb{E}  \log \frac{\exp \left( \beta \log \frac{\pi_\theta(R_{\tau(k)} \mid Q)}{\pi_{\text{ref}}(R_{\tau(k)} \mid Q)} \right)}{\sum_{j=k}^K \exp \left( \beta \log \frac{\pi_\theta(R_{\tau(j)} \mid Q)}{\pi_{\text{ref}}(R_{\tau(j)} \mid Q)} \right)} 
% \end{equation}

\section{Related Work}
\label{sec::related}

\subsection{Alignment of LLMs.}
The language modeling objective of Large Language Models (e.g., predicting the next word) differs from the ultimate goals in LLM applications, such as following instructions and being helpful, factual, and harmless\cite{qi2023fine,bhardwaj2024language,yi2024vulnerability}.
The behavior of pre-trained LLMs may not necessarily align with the principles of their intended use cases. 
Therefore, alignment of LLMs \cite{zhu2024lire,wang2024arithmetic} aims to adjust the outputs of general pre-trained language models to better align with human preferences, significantly improving the performance of LLMs in various downstream applications, such as Summarization\cite{hu2024moments}, dialogue agents \cite{niu2024enhancing}, and question-answering \cite{panda2024holmes}.
Currently, the two most common alignment techniques are instruction tuning \cite{ren2024learning} and reinforcement learning from human feedback (RLHF) \cite{bai2022constitutional,ouyang2022training}. 
Additionally, emerging alignment techniques such as Constitutional AI \cite{bai2022constitutional} and self-alignment \cite{ren2024learning} are also gaining attention.
These primarily focus on embedding alignment rules into pre-trained models to constrain harmful behavior during inference. However, they have not explored how to align objectives with multiple attributes. 
Our study demonstrates that the objectives of preference alignment are influenced by multiple factors.

\subsection{Supervised Alignment}
Large Language Models (LLMs) alignment typically involves two steps. 
The first is supervised fine-tuned (SFT) on high-quality demonstration data to adapt to a specific scenario \cite{stiennon2020learning}.
The second is to learn a strategy for generating high-quality content on preference data to align with human expectations \cite{azar2024general}.
Each preference data item consists of a context, a pair of generated contents, and a pair of human preferences indicating which generated content is better. Additionally, annotating preference data requires some level of expert knowledge.

Learning to align LLMs with human preferences can be achieved through reinforcement learning (RL).
SFT is crucial for ensuring the stable update of the active policy relative to the old policy in preference alignment methods within reinforcement learning \cite{schulman2017proximal}.
In addition, empirical research shows that even in non-RL alignment methods, the SFT is also key to achieve convergence to the desired outcomes \cite{rafailov2024direct, tunstall2023zephyr}.
Therefore, PRO \cite{song2024preference} incorporates the softmax values of the reference response set into the negative log-likelihood loss to merge supervised fine-tuning and preference alignment.
Both SFT and most alignment methods \cite{rafailov2024direct,christiano2017deep,song2024preference,zhao2023slic} rely on annotated data; however, preference data is relatively scarce and expensive to collect in practice \cite{casper2023open}. Therefore, there is an urgent need for an unsupervised method that dynamically annotates preferences during learning to achieve cost-effective preference learning.

\section{Experiments}
%%%%%%%%%%%%%%%%%%%%%%%%%%%%%%%%%%%%%%%%%%%%%%%%%%%%%%%%%%%%%%%%%%%%%%%%%%%%%%%%%%%%%%%%
\begin{algorithm}[t]
\caption{Self-supervised Dynamic Ranking}
\label{algo1}
\KwIn{ \\
$\Delta_{MuAPDF}$: Multi-APDF matrix \\
$ARank$: the order of semantics adopted \(E(Q,R)\) \\
$M$: the size of response \(R = \{R_1, \ldots, R_M\}\) \\}
\KwOut{$DyRank$} 
$DyRank \gets [ \ ]$ \\
\For{$i \gets 0$ \KwTo $M - 1$}{
     $\delta_{max} \gets \max(\Delta_{MuAPDF})$\ ; \\
    $index \gets \text{where}(\Delta_{MuAPDF} == \delta_{max})$\ ; \\
    $row \gets index(0,0)$\ ; 
    $col \gets index(1,0)$\ ; \\
    \If{$ARank(row) < ARank(col)$}{ 
        $DyRank.\text{append}(row)$\ ; \\
        $\Delta_{MuAPDF}(:, row) \gets 0$\ ; \\
        $\Delta_{MuAPDF}(row, :) \gets 0$\ ; 
    } 
    \Else{ 
        $DyRank.\text{append}(col)$\ ; \\
        $\Delta_{MuAPDF}(:, col) \gets 0$\ ; \\
        $\Delta_{MuAPDF}(col, :) \gets 0$\ ; 
    } 
}
$DyRank.\text{append}(ARank.\text{notin}(DyRank))$ \\
\Return $DyRank$\ 
\end{algorithm}
%%%%%%%%%%%%%%%%%%%%%%%%%%%%%%%%%%%%%%%%%%%%%%%%%%%%%%%%%%%%%%%%%%
\begin{table}[h]
\centering
\renewcommand{\arraystretch}{1.1}
\tabcolsep=0.025cm
\begin{tabular}{lcc|lcc}
\toprule
Domain & Volume & RLen & Domain & Volume & RLen \\ 
\midrule
Academia & 16,783 & 4 & Chemistry & 11,058 & 3 \\ 
Cooking & 15,036 & 5 & Electronics & 20,384 & 5 \\ 
History & 6,600 & 3 & Math & 25,860 & 6 \\ 
Music & 16,200 & 4 & Politics & 8,014 & 3 \\ 
Security & 31,327 & 6 & Code & 23,926 & 7 \\ 
\bottomrule
\end{tabular}
\caption{Statistics of the public dataset for Community QA. We align LLMs to QA in different domains, each with varying ranking size (RLen) and data volume.}
\label{domain}
\vspace{-0.2cm}
\end{table}
%%%%%%%%%%%%%%%%%%%%%%%%%%%%%%%%%%%%%%%%%%%%%%%%%%%%%%%%%%%%%%%%%%%%%%%%%%%%%%%%%%%%%%%
\subsection{Baseline} 
Following the DPO \cite{rafailov2024direct}, we evaluated several existing approaches aligned with human preference, including GPT-J \cite{gpt-j} and Pythia-2.8B \cite{biderman2023pythia}.  
Next, we assessed StarCoder2 \cite{lozhkov2024starcoder}, which has demonstrated strong performance in code generation, alongside several general-purpose LLMs: Qwen2 \cite{qwen2}, ChatGLM4 \cite{wang2023cogvlm, glm2024chatglm} and LLaMA serials \cite{touvron2023llama,llama3modelcard}.
Finally, we fine-tuned LLaMA2-7B on the StaCoCoQA and compared its performance with other strong baselines for supervised learning in preference alignment, including SFT, RRHF \cite{yuan2024rrhf}, Silc \cite{zhao2023slic}, DPO, and PRO \cite{song2024preference}.

\begin{table*}[t]
\centering
\renewcommand{\arraystretch}{1.1}
\tabcolsep=0.55cm
\caption{Caption: Statistics of the number of questions with different response pool sizes (Size) in various posting periods (Year) in \(D_I\). Statistics of the number of questions with different response pool sizes (Size) in \(D_Q\) and \(D_C\)}
\begin{tabular}{@{}lcccccc@{}}
\toprule
Year & Size=3 & Size=5 & Size=8 & Size=10 & Size=15 & Size=20 \\ \midrule
Last 2 years & 42,945  & 3,452  & 364    & 148    & 37     & 13     \\
Last 4 years & 178,264 & 18,050 & 2,622  & 1,304  & 408    & 181    \\
Last 6 years & 405,634 & 49,278 & 8,026  & 4,126  & 1,394  & 642    \\
Last 8 years & 719,155 & 100,464 & 18,354 & 9,731  & 3,420  & 1,632  \\ \bottomrule

Step 1 \(D_Q\) & 1,800,588 & 418,688 & 99,646 & 53,681 & 18,429 & 8,513 \\
Step 2 \(D_C\) & 1,428,796 & 311,275 & 69,300 & 37,121 & 12,952 & 6,119 \\ \bottomrule
\end{tabular}
\label{tab:statistics}
\end{table*}

\begin{table*}[h]
\centering
  \renewcommand{\arraystretch}{1.2}
  \tabcolsep=0.44cm
\caption{Statistics on the top 90 categories of StaCoCoQA: Programming Language Categories, Data Volume, and Percentages}
\begin{adjustbox}{width=\linewidth}  
\begin{tabular}{lcc | lcc}
\toprule
\textbf{Category} & \textbf{Volume} & \textbf{Percentage} & \textbf{Category} & \textbf{Volume} & \textbf{Percentage} \\
\midrule
JavaScript & 1,200,942 & 0.120 & Python & 1,028,686 & 0.103 \\
C\# & 741,524 & 0.074 & PHP & 657,849 & 0.066 \\
jQuery & 541,142 & 0.054 & Android & 476,301 & 0.048 \\
CSS & 384,623 & 0.039 & SQL & 341,592 & 0.034 \\
R & 270,346 & 0.027 & Arrays & 247,129 & 0.025 \\
C & 199,767 & 0.020 & ReactJS & 186,690 & 0.019 \\
Node.js & 182,107 & 0.018 & Regex & 169,717 & 0.017 \\
Ruby on Rails & 164,889 & 0.017 & Pandas & 164,879 & 0.017 \\
Python 3.x & 161,735 & 0.016 & SQL Server & 148,887 & 0.015 \\
Swift & 145,214 & 0.015 & ASP.NET & 143,419 & 0.014 \\
.NET & 138,558 & 0.014 & Django & 137,415 & 0.014 \\
Objective-C & 131,735 & 0.013 & Ruby & 122,249 & 0.012 \\
Angular & 120,107 & 0.012 & AngularJS & 119,819 & 0.012 \\
String & 108,758 & 0.011 & Excel & 107,546 & 0.011 \\
XML & 107,448 & 0.011 & TypeScript & 106,706 & 0.011 \\
Ajax & 96,775 & 0.010 & VBA & 90,516 & 0.009 \\
ASP.NET MVC & 88,847 & 0.009 & Bash & 88,632 & 0.009 \\
Laravel & 88,507 & 0.009 & DataFrame & 86,629 & 0.009 \\
Linux & 86,535 & 0.009 & List & 85,043 & 0.009 \\
Spring & 79,137 & 0.008 & WPF & 78,873 & 0.008 \\
PostgreSQL & 78,662 & 0.008 & iPhone & 74,505 & 0.007 \\
MongoDB & 72,507 & 0.007 & Database & 67,669 & 0.007 \\
Oracle & 63,778 & 0.006 & NumPy & 63,055 & 0.006 \\
Multithreading & 61,404 & 0.006 & Scala & 60,979 & 0.006 \\
Function & 60,682 & 0.006 & VB.NET & 59,283 & 0.006 \\
Flutter & 58,351 & 0.006 & & & \\
\bottomrule
\end{tabular}
\end{adjustbox}
\label{tab:stacocoqa_tags}
\end{table*}

\subsection{Implementation Details}
\label{sec::imp}

By limiting the input lengths of \(Q\) and \(R\), and setting thresholds based on the popularity of \(R\), we sampled datasets of various scales from StaCoCoQA: 3K, 8K, 18K, 29K, and 64K, splitting them into training and test sets with a 9:1 ratio. Due to the cost of constructing Gold labels, we selected data from the past four years that are highly popular and feature concise questions as the high-quality test set, totaling 276 samples.
The maximum number of new tokens generated during inference is 128, and beam search decoding is used.
In all following experimental results, PrefHit and PrefRecall correspond to PrefHit@1 and PrefRecall@3, respectively.
We conducted extensive experiments to explore hyperparameters that adapt to datasets of different scales, with varying settings. For detailed information, please refer to the Table~\ref{table::lr}, Table~\ref{table::bs}, Table~\ref{table::w} and Table~\ref{table::scale}.
\begin{table*}[t]
  \renewcommand{\arraystretch}{1.1}
\centering
\caption{Hyperparameter Settings for Training Datasets of Different Scales. The cs represents the convergence step}
  \tabcolsep=0.3cm
  \begin{tabular}{ lcc cccc ccc}
    \toprule
     Scale & batch size &learning rate & evaluation step & epoch & PRO cs & SeAdpra cs  \\ \midrule
        \(Scale=3k\) & 4 & 5e-7 & 200 & 4 &640  & 4,221 \\ 
         \(Scale=8k\) & 4 & 5e-7 & 500 & 3 &2000  &1,000 \\ 
         \(Scale=18k\) & 8 & 5e-7 & 1,000 & 2 & 8000& 2,000 \\ 
         \(Scale=29k\)& 16 & 5e-7& 2,000 & 2 &4000& 6,000\\ 
         \(Scale=64k\) & 32 & 5e-7 & 2,000 & 1&1000 & 3,000\\ 
  \bottomrule
\end{tabular}
\label{table::scale}
\end{table*}

%%%%%%%%%%%%%%%%%%%%%%%%%%%%%%%%%%%%%%%%%%%%%%%%%%%%%%%%%%%%%%%%%%%%%%%%%%%%%%%%%%%%%%%%%%%
\begin{table*}[t]
\centering
\caption{Results of experiments with different weight \(\alpha\) in Perceptual Alignment.}
\label{alpha}
\renewcommand{\arraystretch}{1.13} % Adjust row height
\tabcolsep=0.15cm % Adjust column spacing
\begin{tabular}{lccccccccc}
\toprule
\multirow{2}{*}{Method} & \multicolumn{6}{c}{Preference \((\uparrow)\)} & \multicolumn{3}{c}{Accuracy \((\uparrow)\)} \\ 
\cmidrule(lr){2-7} \cmidrule(lr){8-10}
& \small{PrefHit@1} & \small{PrefHit@3} & \small{PrefRec@2} & \small{PrefRec@4} & \small{Reward1} & \small{Reward2} 
& \small{CodeSim} & \small{BLEU} & \small{RougeL} \\ 
\midrule
\( \alpha=0.01 \) & 0.3659 & 0.5326 & 0.5036 & 0.8279 & 0.2301 & 0.8233 & 0.6900 & 0.1412 & 0.2078 \\
\(\alpha=0.05\)  & 0.3478 & 0.5471 & 0.5127 & 0.8252 & 0.2233 & 0.8405 & 0.6914 & 0.1741 & 0.2182 \\
\(\alpha=0.1\)   & 0.3225 & 0.5072 & 0.4819 & 0.8315 & 0.2311 & 0.8320 & 0.6901 & 0.2177 & 0.1557 \\  
\( \alpha=0.2 \) & 0.3370 & 0.5254 & 0.4964 & 0.8297 & 0.2304 & 0.8212 & 0.6896 & 0.1352 & 0.2080 \\
\( \alpha=0.5 \)  & 0.2826 & 0.4819 & 0.4565 & 0.8179 & 0.1901 & 0.7612 & 0.6752 & 0.1013 & 0.1654 \\
\( \alpha=1 \)   & 0.3225 & 0.5145 & 0.4891 & 0.8342 & 0.2241 & 0.8330 & 0.6901 & 0.1534 & 0.2168 \\
\bottomrule
\end{tabular}
\label{table::w}
\end{table*}

%%%%%%%%%%%%%%%%%%%%%%%%%%%%%%%%%%%%%%%%%%%%%%%%%%%%%%%%%%%%%%%%%%%%%%%%%%%%%%%%%%%%%%%%%%%%%%%%%%%%%%%

\begin{table*}[h]
\setlength{\tabcolsep}{1.5pt}
\centering
 \caption{Results of experiments on the different sizes of response \(Step\). }
\label{size}
\renewcommand{\arraystretch}{1.13}
   \tabcolsep=0.15cm
% \resizebox{\textwidth}{3cm}{
\begin{tabular}{ lc ccc ccccc}
\toprule
\multirow{2}{*}{Method}&\multicolumn{6}{c}{Preference \((\uparrow)\)} & \multicolumn{3}{c}{Accuracy \((\uparrow)\)} \\ \Xcline{2-7 }{0.4pt}  \Xcline{ 8-10}{0.4pt} 
 & \small{PrefHit@1} & \small{PrefHit@3} & \small{PrefRec@2} & \small{PrefRec@4} &\small{Reward1} &\small{Reward2}&\small{CodeSim}  & \small{BLEU}&\small{RougeL}   \\ \midrule
      \( Step=2\) & 0.3333 & 0.5217 & 0.5000 & 0.8279 & 0.2347 & 0.8226 & 0.6902 & 0.2081 & 0.1436 \\
      
          \( Step=3\) & 0.3370 & 0.5217 & 0.4891 & 0.8270 & 0.2339 & 0.8219 & 0.6904 & 0.2085 & 0.1420 \\
      
          \( Step=4\) & 0.3261 & 0.5145 & 0.4801 & 0.8252 & 0.2309 & 0.8136 & 0.6881 & 0.2065 &  0.1432 \\
       
          \( Step=5\) & 0.3261 & 0.5109 & 0.4873 & 0.8388 & 0.2245 & 0.8307 & 0.6898 & 0.2172 & 0.1548 \\
  \bottomrule
\end{tabular}
\label{table::step}
\end{table*}
%%%%%%%%%%%%%%%%%%%%%%%%%%%%%%%%%%%%%%%%%%%%%%%%%%%%%%%%%%%%%%%%%%%%%%%%%%%%%%%%%%%%%%%%
\begin{table*}[h]
\setlength{\tabcolsep}{2pt}
\centering
 \caption{Results of experiments on the different learning rate \(lr\). }
\label{lr}
\renewcommand{\arraystretch}{1.13}
   \tabcolsep=0.15cm
% \resizebox{\textwidth}{3cm}{
\begin{tabular}{ lc ccc ccccc}
\toprule
\multirow{2}{*}{Method}&\multicolumn{6}{c}{Preference \((\uparrow)\)} & \multicolumn{3}{c}{Accuracy \((\uparrow)\)} \\ \Xcline{2-7 }{0.4pt}  \Xcline{ 8-10}{0.4pt} 
 & \small{PrefHit@1} & \small{PrefHit@3} & \small{PrefRec@2} & \small{PrefRec@4} &\small{Reward1} &\small{Reward2}&\small{CodeSim}  & \small{BLEU}&\small{RougeL}   \\ \midrule
          \( lr=1e-7\) & 0.3333 & 0.5217 & 0.5000 & 0.8279 & 0.2347 & 0.8226 & 0.6902 & 0.2081 & 0.1436 \\
      
          \( lr=3e-7\) & 0.3370 & 0.5217 & 0.4891 & 0.8270 & 0.2339 & 0.8219 & 0.6904 & 0.2085 & 0.1420 \\
      
          \( lr=5e-7\) & 0.3478 & 0.5471 & 0.5127 & 0.8252 & 0.2233 & 0.8405 & 0.6914 & 0.1741 & 0.2182 \\
       
          \( lr=1e-6\) & 0.2899 & 0.4891 & 0.4692 & 0.8297 & 0.2322 & 0.8082 & 0.6872 & 0.1330 & 0.2056 \\
          \( lr=5e-6\) & 0.3080 & 0.5471 & 0.4964 & 0.8234 & 0.2156 & 0.8465 & 0.6945 & 0.1742 & 0.2274 \\
          \( lr=1e-5\) & 0.3261 & 0.5109 & 0.4783 & 0.8225 & 0.2021 & 0.8494 & 0.6971 & 0.1955 & 0.2216 \\
  \bottomrule
\end{tabular}
\label{table::lr}
\end{table*}
%%%%%%%%%%%%%%%%%%%%%%%%%%%%%%%%%%%%%%%%%%%%%%%%%%%%%%%%%%%%%%%

% \begin{table*}[h]
% \centering
% \caption{Results of experiments on the different training data scale \(Scale\).}
%   \renewcommand{\arraystretch}{1.1}
%   \tabcolsep=0.15cm
%   \begin{tabular}{ lcc cccc ccc}
%     \toprule
% \multirow{2}{*}{Method}&\multicolumn{6}{c}{Preference \((\uparrow)\)} & \multicolumn{3}{c}{Accuracy \((\uparrow)\)} \\ \Xcline{2-7 }{0.4pt}  \Xcline{ 8-10}{0.4pt} 
%  & \small{PrefHit@1} & \small{PrefHit@3} & \small{PrefRec@2} & \small{PrefRec@4} &\small{Reward1} &\small{Reward2}&\small{CodeSim}  & \small{BLEU}&\small{RougeL}   \\ \midrule

%         \(Scale=3k\) & 32.61 & 53.26 & 48.91 & 82.43 & 69.28 & 26.32 &26.45& 16.31 \\ 
%          \(Scale=8k\) & 25.72 & 42.49 & 41.30 & 81.25 & 68.63 & 24.90&23.33 & 11.41 \\ 
%          \(Scale=18k\) & 25.72 & 42.49 & 41.30 & 81.25 & 68.63 & 24.90&23.33 & 11.41 \\ 

%          \(Scale=29k\)& - & -& 41.30 & 81.25 & 68.63 & 24.90&23.33 & 11.41 \\ 
%          \(Scale=64k\) & - & 42.49 & 41.30 & 81.25 & 68.63 & 24.90&23.33 & 11.41 \\ 
%   \bottomrule
% \end{tabular}
% \end{table*}
%%%%%%%%%%%%%%%%%%%%%%%%%%%%%%%%%%%%%%%%%%%%%%%%%%%%%%%%%%%%%%%%%%%%%%%%%%%%%%%%%%%%%%%%%%%%%%%%%%%%%
\begin{table*}[t]
\setlength{\tabcolsep}{3pt}
\centering
 \caption{Results of experiments on the different batch sizes \(size\) during training. }
\label{bs}
\renewcommand{\arraystretch}{1.13}
   \tabcolsep=0.15cm
% \resizebox{\textwidth}{3cm}{
\begin{tabular}{ lc ccc ccccc}
\toprule
\multirow{2}{*}{Method}&\multicolumn{6}{c}{Preference \((\uparrow)\)} & \multicolumn{3}{c}{Accuracy \((\uparrow)\)} \\ \Xcline{2-7 }{0.4pt}  \Xcline{ 8-10}{0.4pt} 
 & \small{PrefHit@1} & \small{PrefHit@3} & \small{PrefRec@2} & \small{PrefRec@4} &\small{Reward1} &\small{Reward2}&\small{CodeSim}  & \small{BLEU}&\small{RougeL}   \\ \midrule
       \( size=4\) & 0.3659 & 0.5326 & 0.5036 & 0.8279 & 0.2301 & 0.8233 & 0.6900 & 0.2079 & 0.1412 \\
        \( size=8\)& 0.3261 & 0.5471 & 0.5072 & 0.8225 & 0.2220 & 0.8369 & 0.6903 & 0.1603 & 0.2159 \\

         \(size=16\)& 0.3514 & 0.5326 & 0.4946 & 0.8225 & 0.2392 & 0.8294 & 0.6911 & 0.1571 & 0.2160 \\ 
       
        \(size=32\) & 0.2609 & 0.4275 & 0.4094 & 0.8107 & 0.4454 & 0.7396 & 0.6856 & 0.1326 & 0.1330 \\ 
      
        \(size=64\) & 0.2572 & 0.4384 & 0.4130 & 0.8116 & 0.4595 & 0.7448 & 0.6860 & 0.1372 & 0.1374 \\ 
       
        \(size=128\) & 0.2428 & 0.4167 & 0.4185 & 0.8125 & 0.4738 & 0.7464 & 0.6862 & 0.1364 & 0.1370 \\ 
       
  \bottomrule
\end{tabular}
\label{table::bs}
\end{table*}
%%%%%%%%%%%%%%%%%%%%%%%%%%%%%%%%%%%%%%%%%%%%%%%%%%%%%%%%%%%%%%%%%%%%%%%%%%%%%%%%%%%%%%%%%%%%%%%%%%%%%%%%%%
\begin{figure*}[t]
    \centering
    \includegraphics[width=\linewidth]{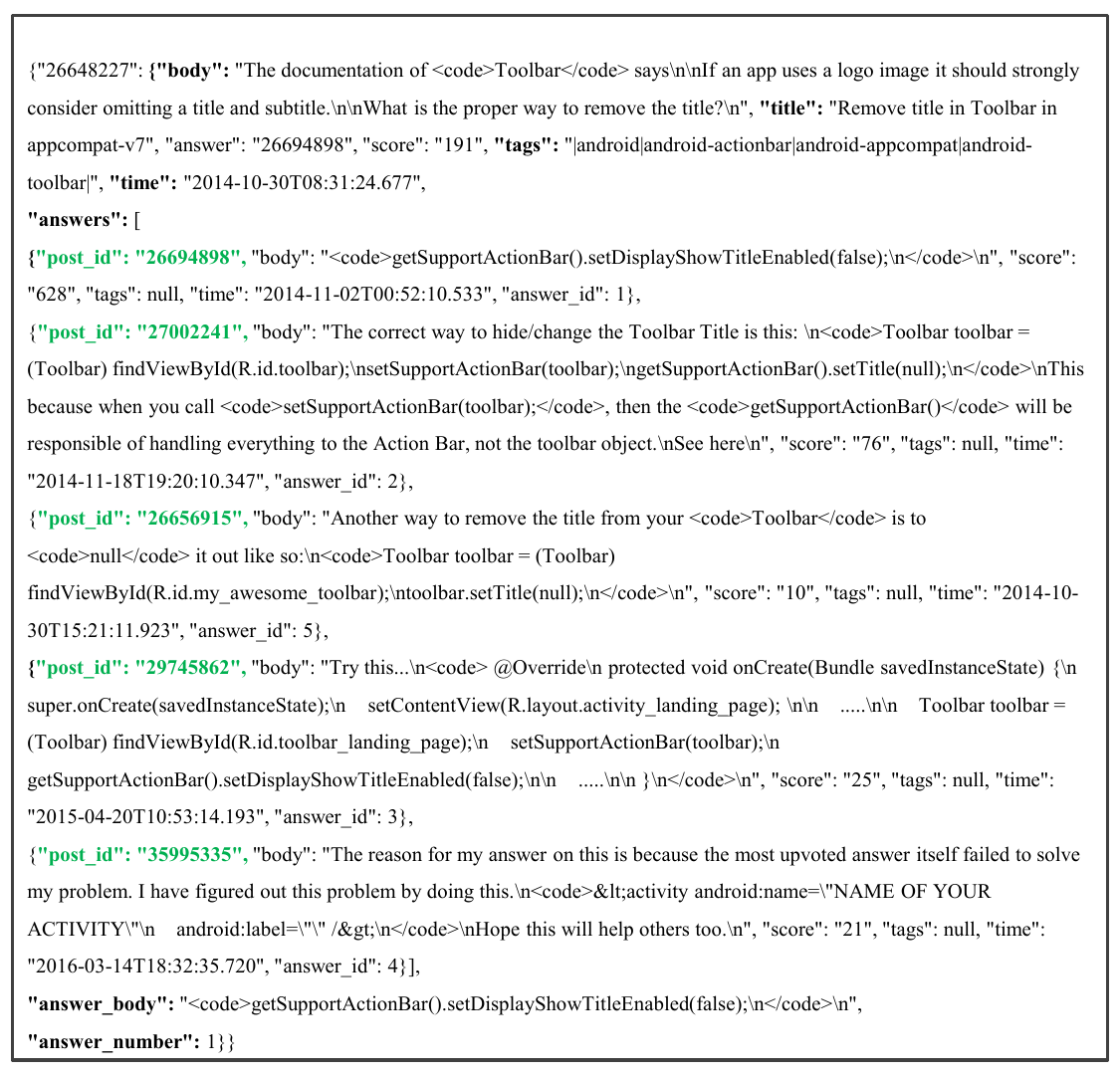}
    \caption{An example from the our proposed programming dataset StaCoCoQA.}
    \label{fig::stacocoqa}
\end{figure*}
\begin{figure*}[t]
    \centering
    \includegraphics[width=\linewidth]{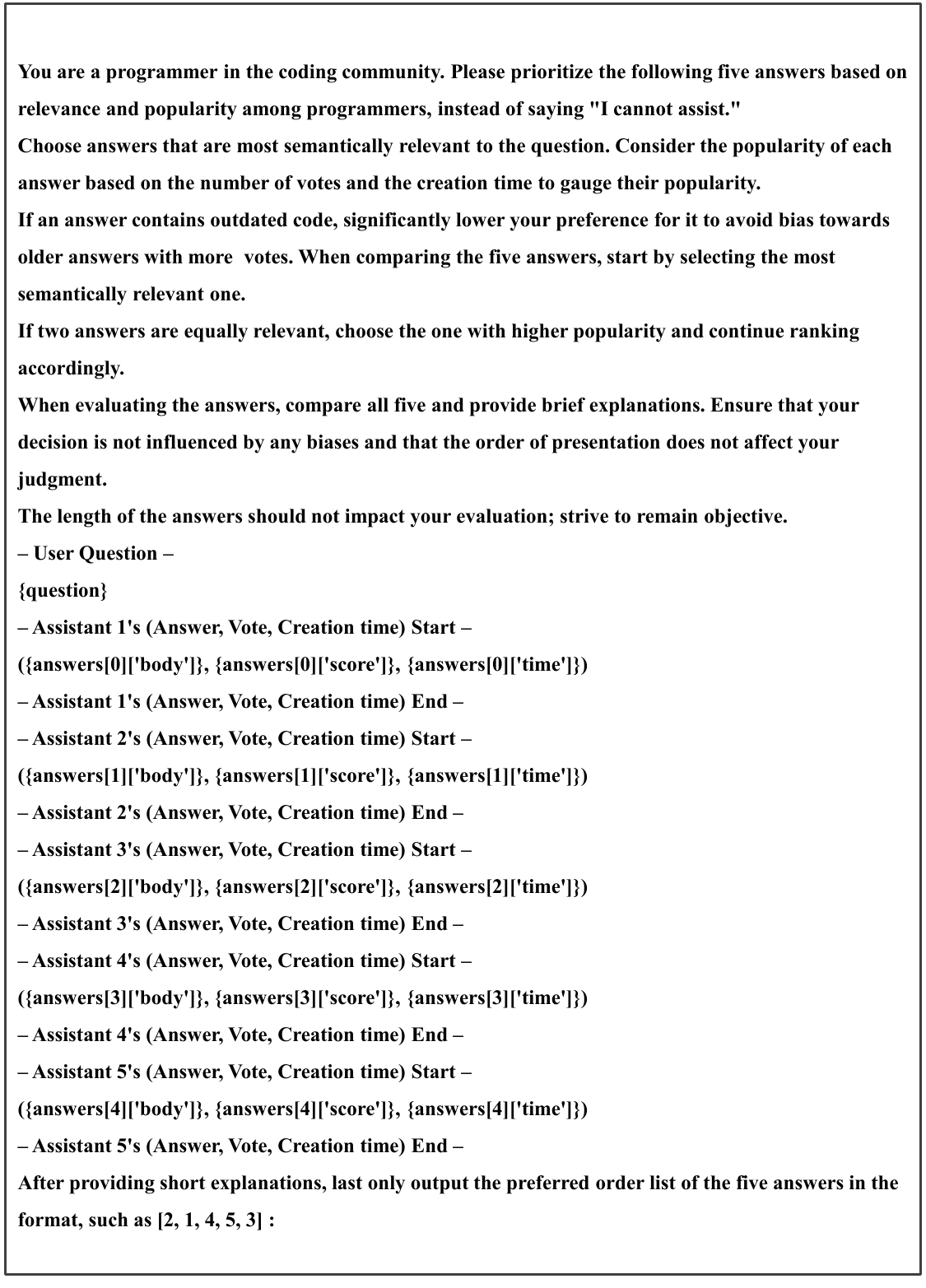}
    \caption{Rules for labeling StaCoCoQA testing data, whether manually or AI-assisted, consider semantic relevance, popularity, and creation time, with a time-decay adjustment applied to popularity.}
    \label{fig::gpt4}
\end{figure*}
\begin{figure*}[t]
    \centering
    \includegraphics[width=\linewidth]{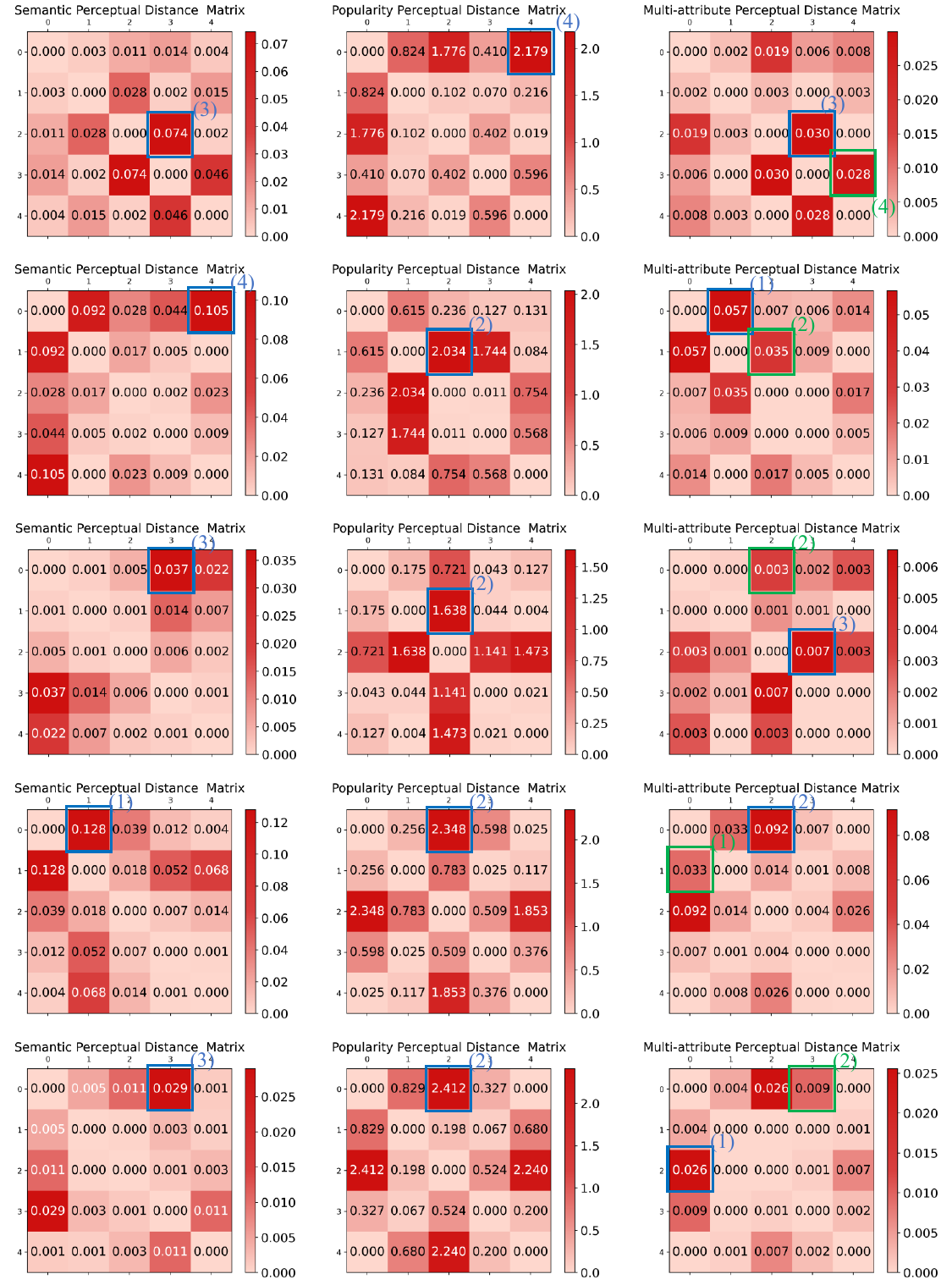}
    \caption{The visualization of Attribute-Perceptual Distance Factors (APDF) for diifferent selected samples having five candidates. The blue represents the alignment target of the corresponding APDF. The green indicates that the second alignment target is suboptimal compared to the blue one. We have three key findings: (1) The alignment of the Multi-attribute Perceptual Distance Matrix \(\Delta_{M}\) could be the alignment target of the Semantic Perceptual Distance Matrix \(\Delta_{Se}\). (2) The alignment target of the \(\Delta_{M}\) could also be the alignment target of the Popularity Perceptual Distance Matrix \(\Delta_{Po}\). (3) The alignment target of the \(\Delta_{M}\) may neither be the alignment target of the \(\Delta_{Se}\) nor the \(\Delta_{Po}\).}
    \label{fig::hot1}
\end{figure*}
\begin{figure*}[t]
    \centering
    \includegraphics[width=\linewidth]{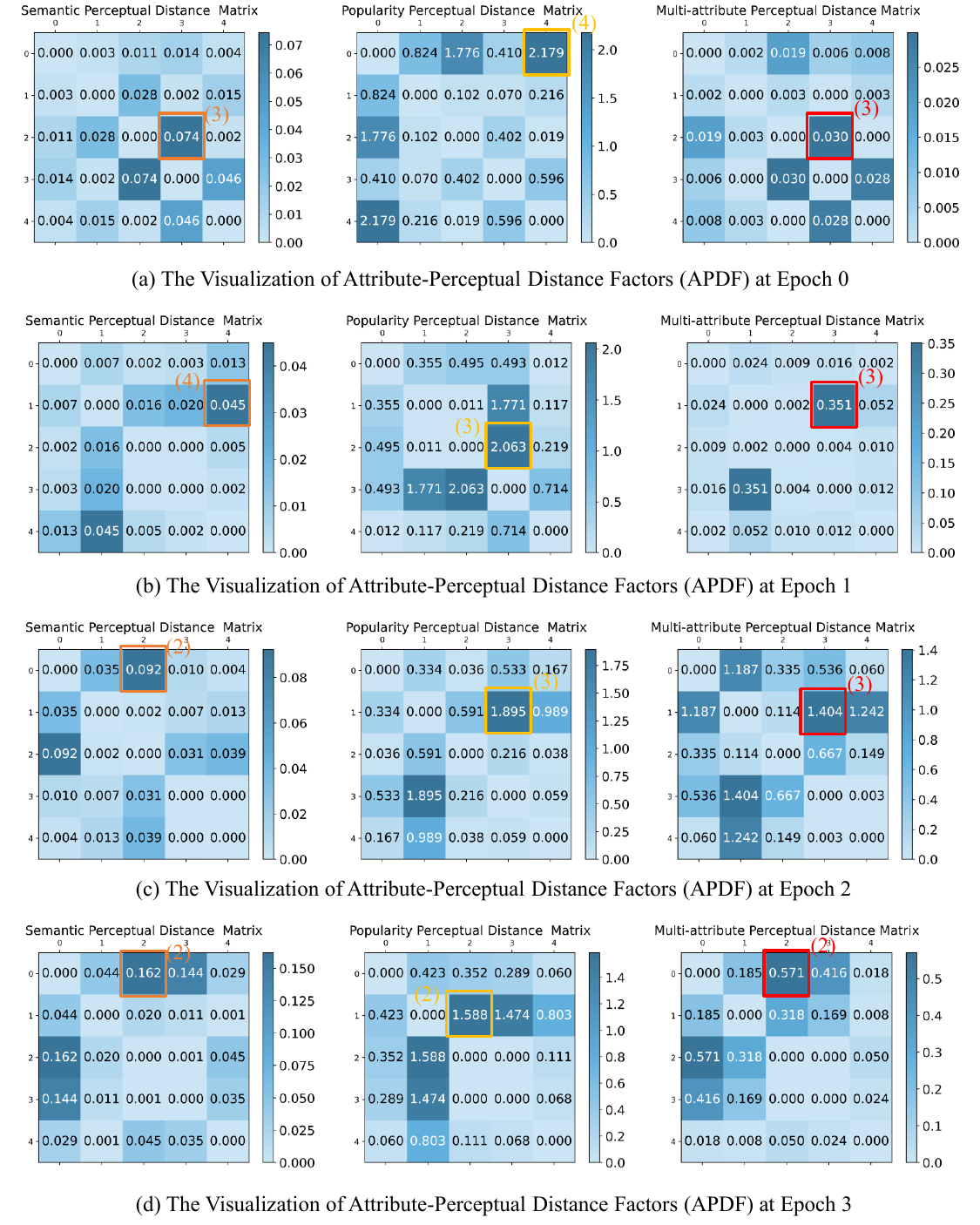}
    \caption{Visualization of the alignment target evolution for a sample throughout the training process. The orange represents the alignment target of the Semantic Perceptual Distance Matrix \(\Delta_{Se}\). The yellow represents the alignment target of the Popularity Perceptual Distance Matrix \(\Delta_{Po}\). The red represents the alignment target of the Multi-attribute Perceptual Distance Matrix \(\Delta_{M}\). We have two key findings. (1) At the same epoch, the alignment targets may differ across the Semantic Perceptual Distance Matrix \(\Delta_{Se}\), the Popularity Perceptual Distance Matrix \(\Delta_{Po}\), and the Multi-attribute Perceptual Distance Matrix \(\Delta_{M}\). (2) Across different epochs, the alignment targets for the same Attribute-Perceptual Distance Matrix may evolve.}
    \label{fig::hot2}
\end{figure*}

\label{sec:appendix}

\end{document}